%% file: main.tex
\documentclass[acmsmall,natbib=true,screen]{acmart}
\usepackage{balance}
\usepackage{tabularx, booktabs}
\usepackage{multirow}
\usepackage{makecell}
\usepackage{threeparttable}
\usepackage{subcaption}
\usepackage{tablefootnote}
\usepackage{geometry}
\usepackage{array}
\usepackage[ruled, linesnumbered]{algorithm2e}
\usepackage{xcolor}
\usepackage{subcaption}
\usepackage{textcomp}
\definecolor{BLUE}{rgb}{0,0,200}
\definecolor{blue}{rgb}{0,0,0}
\definecolor{largeblue}{RGB}{86,108,215}
\definecolor{lightblue}{RGB}{173,216,230} 
\definecolor{trueblue}{rgb}{0,0,200}
\definecolor{myPurple}{RGB}{128,0,128}
\definecolor{myLightPurple}{RGB}{200,150,200}
\SetKw{kwInit}{Init}

\AtBeginDocument{%
  }

\setcopyright{acmcopyright}
\copyrightyear{2023}
\acmYear{2023}
\acmDOI{XXXXXXX.XXXXXXX}

\acmJournal{TOIS}
\acmVolume{37}
\acmNumber{4}
\acmArticle{111}
\acmMonth{12}

\begin{document}
\title{Relevance Feedback with Brain Signals}

\author{Ziyi Ye}
\email{yeziyi1998@gmail.com}
\affiliation{%
  \institution{Quan Cheng Lab, DCST, Tsinghua University}
  \city{Beijing}
  \country{China}
}
\author{Xiaohui Xie}
\email{xiexh_thu@163.com}
\affiliation{%
  \institution{Quan Cheng Lab, DCST, Tsinghua University}
  \city{Beijing}
  \country{China}
}
\author{Qingyao Ai}
\email{aiqy@tsinghua.edu.cn}
\affiliation{%
  \institution{Quan Cheng Lab, DCST, Tsinghua University}
  \country{China}
}
\author{Yiqun Liu}
\email{yiqunliu@tsinghua.edu.cn}
\affiliation{%
  \institution{Quan Cheng Lab, DCST, Tsinghua University}
  \city{Beijing}
  \country{China}
}
\author{Zhihong Wang}
\email{wangzhh629@mail.tsinghua.edu.cn}
\affiliation{%
  \institution{Quan Cheng Lab, DCST, Tsinghua University}
  \city{Beijing}
  \country{China}
}
\author{Weihang Su}
\email{swh22@mails.tsinghua.edu.cn}
\affiliation{%
  \institution{Quan Cheng Lab, DCST, Tsinghua University}
  \city{Beijing}
  \country{China}
}
\author{Min Zhang}
\email{z-m@tsinghua.edu.cn}
\affiliation{%
  \institution{Quan Cheng Lab, DCST, Tsinghua University}
  \city{Beijing}
  \country{China}
}

\renewcommand{\shortauthors}{Ye et al.}

\begin{abstract}
The Relevance Feedback~(RF) process relies on accurate and real-time relevance estimation of feedback documents to improve retrieval performance.
Since collecting explicit relevance annotations imposes an extra burden on the user, extensive studies have explored using pseudo-relevance signals and implicit feedback signals as substitutes.
However, such signals are indirect indicators of relevance and suffer from complex search scenarios where user interactions are absent or biased.

Recently, the advances in portable and high-precision brain-computer interface~(BCI) devices have shown the possibility to monitor user's brain activities during search process. 
Brain signals can directly reflect user's psychological responses to search results and thus it can act as additional and unbiased RF signals. 
To explore the effectiveness of brain signals in the context of RF, we propose a novel RF framework that combines BCI-based relevance feedback with pseudo-relevance signals and implicit signals to improve the performance of document re-ranking. 
The experimental results on the user study dataset show that incorporating brain signals leads to significant performance improvement in our RF framework. 
Besides, we observe that brain signals perform particularly well in several hard search scenarios, especially when implicit signals as feedback are missing or noisy. This reveals when and how to exploit brain signals in the context of RF.

\end{abstract}




\maketitle

\input{1_introduction.tex}
\input{2_related_work.tex}

\input{3_data_collection.tex}

\input{4_method.tex}

\input{5_experiments_results}

\input{6_conclusions.tex}

\input{7_acknowledgement.tex}

\bibliographystyle{ACM-Reference-Format}

\balance
\bibliography{references}
\appendix
\newpage

\input{8_appendix.tex}
\end{document}

%% file: 1_introduction.tex
\section{introduction}
\textcolor{blue}{The application of Relevance Feedback~(RF) is crucial to elicit additional information beyond the initial query, since queries submitted to the search engine are usually short, vague, and sometimes ambiguous~\cite{4azad2019query}.}
RF~\cite{26harman1992relevance,63white2002use} method is widely applied to improve retrieval accuracy by acquiring feedback information in addition to the submitted query. 
In a standard RF process, users should provide explicit relevance judgments on a given collection of documents. 
The search system then automatically re-ranks the search results with information extracted from the documents annotated as relevant or irrelevant by the user.
RF paradigms effectively improve retrieval performance~\cite{26harman1992relevance,47montazeralghaem2020reinforcement,54pereira2020iterative,63white2002use}, especially in the cases where users do not have a thorough idea of what information they are searching for or have difficulty transforming their search intents into queries~\cite{59ruthven2003survey}.

\textcolor{blue}{Given that the standard RF process frequently places an additional cognitive burden on users, as it necessitates explicit relevance indications, prior research has explored a range of alternative signals to estimate the relevance of search results.}
These signals can be broadly categorized into two groups: pseudo-relevance signals and implicit signals.
The basic idea of pseudo-relevance signals is simply treating the top-ranked documents as relevant and utilizing their terms for query rewriting, which is often applied to alleviate the problem of term mismatching~\cite{38lavrenko2017relevance,42lin2021multi}. 
Pseudo-relevance signals do not necessarily require any user signals, making them easy to deploy and widely used. 
However, the quality of pseudo-relevance signals highly depends on the effectiveness of the initial retrieval~\cite{39li2022does} and is particularly unstable when the submitted query is vague or ambiguous.

Implicit signals~(e.g., click, dwell time), on the other hand, are usually considered more reliable than pseudo-relevance signals, especially in interactive Information Retrieval~(IR) scenarios~\cite{8bi2019iterative,54pereira2020iterative}. 
Nevertheless, implicit signals are indirect probes of relevance inferred from user’s behaviors and thus are often biased and inaccurate~\cite{44liu2014skimming}. 
\textcolor{blue}{
While existing research attempts to mitigate these biases by utilizing additional signals such as eye-tracking and mouse movement, practical experience has revealed that these indirect signals still bring biases. 
For instance, the credibility of eye-tracking signals is limited as gazing at a document can not necessarily guarantee that its content is genuinely relevant~\cite{mao2014estimating}.
}
Due to the inaccuracy and inheriting bias of pseudo-relevance/implicit signals, the effectiveness of RF is often limited~\cite{39li2022does}. 
Hence, the potential advancement of RF lies in acquiring more accurate and unbiased signals.

Recently, the emergence of neurological techniques has attracted researchers to explore interactive information systems with brain signals~\cite{18davis2021collaborative,71ye2022brain}. 
As neurological devices become portable and affordable, it is possible to capture user’s brain signals in realtime and build a practical search system with brain–computer interface~(BCI)~\cite{11chen2022web}. 
Beyond applying BCI to control search engines, existing studies~\cite{56pinkosova2020cortical,72ye2022don} have revealed underlying differences between brain responses to relevant and irrelevant search results. 
These differences indicate that brain signals can be acquired as novel alternative signals to measure relevance.
\textcolor{blue}{In contrast to existing signals, brain signals directly reflect a user's psychological activities and, as a result, are less susceptible to the inherent biases associated with user behaviors.
Yet, to the best of our knowledge, the potential benefits of brain signals in improving RF performance remain mostly unknown, especially when combined with existing signals.
}

In this paper, we propose a novel RF framework combining brain signals with pseudo-relevance signals and click signals to improve RF performance. 
The proposed framework can be employed in two distinct RF settings, i.e., iterative RF~(IRF) and retrospective RF~(RRF). 
IRF~\cite{1aalbersberg1992incremental} dynamically re-ranks upcoming documents as more search results are presented to the user, which is preferable in situations where user signals are collected incrementally~\cite{8bi2019iterative}. 
We are interested in analyzing how brain signals could improve the quality of RF while the search is ongoing. 
In contrast, the retrospective RF~(RRF) setting focuses on re-estimating the relevance of historically examined documents after the search process ends. 
Retrospective RF~(RRF) cannot directly facilitate the ongoing search process, but it can help the potential search process with a similar search intent for other users~\cite{31joachims2017unbiased}. 
\textcolor{blue}{In both IRF and RRF, our objective is to investigate whether brain signals can function as unbiased indicators of relevance. 
Additionally, we examine search scenarios where conventional user signals are biased, highlighting potential applications for BCI within IR.
}

To verify the effectiveness of IRF and RRF with brain signals, we conduct a lab-based Web search study and collect corresponding pseudo-relevance signals, click signals, and brain signals during the search process. Based on the collected dataset, we conduct document re-ranking tasks in the context of IRF and RRF, respectively. 
With extensive experiments, we demonstrate the effectiveness of the proposed framework, especially the benefits brought by brain signals. 
We observe that brain signals lead to an additional improvement of 8.8\% and 7.4\% in terms of NDCG@10~(Normalized Discounted Cumulative Gain) for IRF and RRF, respectively. 
Furthermore, we delve into search scenarios in which brain signals are more effective than pseudo-relevance and click signals. 
\textcolor{blue}{
We observe that brain signals are particularly helpful in the cases where ``bad click'' happens, which may alleviate the ``clickbait'' issue~(i.e., documents with misleading headlines attract bad click~\cite{45lu2018between}).
} 
\textcolor{blue}{Besides, we have observed that brain signals can offer valuable relevance guidance for extracting information from non-clicks, a particularly advantageous feature for initiating RF before any click occurs.}
Drawing from these findings, we propose using a method that can boost RF performance by adaptively adjusting the combination weight of brain signals and other signals based on specific search scenarios.

In summary, our contributions are three-fold. 
(1)~We devise a novel RF framework that combines pseudo-relevance signals, click signals, and brain signals to improve the performance of document re-ranking tasks. 
(2)~We conduct a user study in Web search scenario to explore the improvement brought by brain signals in the proposed framework. 
(3)~We investigate the possible search scenarios in which brain signals are more effective than existing signals. This reveals when and how we can bring the benefit of BCI into search systems.

The rest of this article is organized as follows. 
In the next section, we review related work in relevance feedback and neuroscientific approach. 
In Section~\ref{3}, we introduce the Web search study and the data collection procedures. 
Then in Section~\ref{4}, the preliminaries about the RF tasks~(IRF and RRF) and the proposed RF framework are elaborated. 
Next, we present the experimental results and corresponding analyses to explore the impact of brain signals in the context of RF in Section~\ref{5}. 
Finally, we conclude this work and discuss its applications and limitations in Section~\ref{6}.

%% file: 2_related_work.tex
\section{related work}
\label{2}
\subsection{Relevance Feedback Signals}
The standard RF process was based on explicit signals that would impose additional manual efforts and hurt the search experience. 
Hence, pseudo-relevance signals~\cite{42lin2021multi,60wang2021pseudo} were often applied by simply assuming the top-ranked documents as relevant. 
However, \citet{39li2022does} observed that the performance of pseudo-relevance signals is often limited if the initial retrieval performance is weak. 
In addition to pseudo-relevance signals, implicit signals were also widely studied, in which a user’s behaviors~(e.g., click~\cite{63white2002use}, dwell time~\cite{48morita1994information}, and eye-tracking~\cite{2akuma2022eye}) are used to infer their preference. Among all the implicit signals, click signals were the most widely used and were thought to be an indication of relevance~\cite{63white2002use}. 
However, click signals are indirect probes of relevance~\cite{44liu2014skimming} and may be biased in several search scenarios~\cite{20dolma2021improving,40li2009good,65wu2020credibility}.

Traditionally, RF is based on only one type of signals~(e.g., pseudo-relevance signals~\cite{42lin2021multi,60wang2021pseudo,74yu2021improving} or implicit signals~\cite{15claypool2001implicit,62white2006implicit}), which limited their performance due to the inaccuracy and inheriting bias of existing RF signals~\cite{39li2022does}. 
To address this biased problem, prior literature has devised several robust architectures~\cite{10can2014incorporating,46lv2009adaptive,73yin2014temporal} to deal with biased signals. 
For example, \citet{46lv2009adaptive} proposed to learn an adaptive coefficient that can avoid overvaluing RF information when RF signals may be unreliable. 
Besides, there existed a series of studies on click models~\cite{12chuklin2022click,77zhang2021constructing} have been conducted in an effort to estimate relevance from biased click signals.

Nevertheless, such approaches are still limited since user interactions are extremely varied and hard to probe~\cite{75zhang2020user}. 
Hence, potential advancement lies in acquiring more accurate RF signals. 
This paper explores brain signals, which directly capture the user’s thoughts and thus are unbiased and accurate. 
With empirical experiments, we verify that brain signals can bring additional improvement on top of existing signals, especially in situations where implicit signals are biased.

\subsection{Relevance Feedback Techniques}
Standard RF techniques required relevance assessments on a fixed batch of documents. Since pseudo-relevance signals are the most commonly used signals in standard RF, this kind of RF is also named as top-k RF. There were two main streams of top-k RF techniques, i.e, vector space model~(VSM)- based~(e.g., Rocchio~\cite{57rocchio1971relevance}) and language model~(LM)-based (e.g., RM3~\cite{38lavrenko2017relevance}). The VSM-based method refines the query vector to be closer to the relevant documents~\cite{57rocchio1971relevance}, while the LM-based method expands the query by selecting relevant terms~\cite{38lavrenko2017relevance}. Recently, with the developments of neural retrieval methods~\cite{53nogueira2019passage,69yang2019end}, researchers have attempted to explore the effectiveness of RF with BERT-based ranker~\cite{42lin2021multi,60wang2021pseudo,61wang2023colbert,74yu2021improving}. For example, \citet{60wang2021pseudo,78zheng2020bert}adopted a VSM-based design to fuse BERT-based query embeddings with document embeddings. 
\citet{78zheng2020bert} proposed BERT-QE which treats the top-ranked documents as additional queries to retrieve documents with a BERT-based re-ranker.

In addition to top-k RF, alternative RF techniques are devised in different settings. 
For example, \citet{1aalbersberg1992incremental} propose iterative RF~(IRF), which re-ranks upcoming documents after each user interaction in an incremental manner. 
This approach is particularly advantageous in practical search processes, especially when user signals are collected incrementally [8]. 
Besides transforming top-k RF techniques into IRF settings~\cite{1aalbersberg1992incremental}, recent works also devise novel IRF techniques for various scenarios, e.g., conversational search~\cite{8bi2019iterative} and product search~\cite{9bi2019conversational}. 
In this paper, we first explore IRF as brain signals can provide real-time guidance for RF. 
In addition to up-coming documents, we also supplement the relevance re-estimation for historical documents, namely retrospective RF~(RRF) in this paper. 
RRF helps potential search processes with similar intents in the future by directly re-ranking the search results or constructing learning-to-rank models~\cite{31joachims2017unbiased} with the estimated relevance signals. 
Conventionally, RRF relies on probing user behaviors~(especially clicks) and strongly suffers from biased user behaviors~\cite{12chuklin2022click,31joachims2017unbiased}. 
Hence, exploring alternative unbiased signals, i.e., brain signals in RRF is valuable. The construction of IRF and RRF tasks are elaborated in Section~\ref{4.1}.

\subsection{Neuroscientific Approach for IR}
\label{2.3}
There is increasing literature that adopts neuroscientific methods into IR scenarios~\cite{3allegretti2015relevance,11chen2022web,23eugster2014predicting,25gwizdka2017temporal,52mostafa2016deepening}. 
These studies involved a variety of neuroimaging techniques, such as functional magnetic resonance imaging~(fMRI)~\cite{50moshfeghi2018search,51moshfeghi2016understanding}, magnetoencephalogram~(MEG)~\cite{32kauppi2015towards}, and electroencephalogram~(EEG)~\cite{18davis2021collaborative,56pinkosova2020cortical,72ye2022don}. 
Among these techniques, fMRI has the highest spatial resolution, which can help identify the topological distribution of cognitive functions during information retrieval. 
For example, \citet{51moshfeghi2016understanding} examined the emergence of Information Need~(IN) and identified activated brain regions. 
Their further study predicted the realization of an IN using fMRI voxels from a generalized set of brain regions~(GM) and a unique set of regions for each individual~(PM), respectively. 
Compared to fMRI, MEG and EEG have higher temporal resolutions, which can help us understand the temporal dynamics of brain activity and construct real-time BCI applications. 
MEG produces better spatial resolution than EEG, but it is rather expensive and has strict requirements on the experimental environment where external magnetic signals should be shielded. 
On the other hand, EEG is more frequently used to construct real-time BCI applications due to its portability and affordability~\cite{33kawala2021summary,36kohli2022review}. 
For example, \citet{11chen2022web} propose an EEG-based demonstration search system that can obtain dictates in the steady-state visual evoked potential~(SSVEP) for controlling search engines. 
\citet{3allegretti2015relevance} studied the time frame to distinguish human’s brain response to relevant and irrelevant items. 
\textcolor{blue}{In their subsequent research~\cite{56pinkosova2020cortical}, the graded phenomenon of relevance was further investigated, i.e., a search item could be partially relevant to the user.}

Among these works, a common finding demonstrated by \citet{3allegretti2015relevance,23eugster2014predicting} is that brain signals can play as an indicator of relevance. 
For example, \citet{23eugster2014predicting} adopted brain signals to predict term relevance, and their further research utilized this prediction to interactively recommend other relevant information~\cite{22eugster2016natural}. 
\citet{3allegretti2015relevance,24golenia2018implicit}, on the other hand, predicts user’s preference to image stimuli with brain signals. 
A more recent work~\citet{72ye2022don} explored relevance in a more practical Web search scenario and observed that brain signals could detect the phenomenon of ``good abandonment''. 
However, their studies lacked applications for Web retrieval and limited user interactions~(e.g., allowing interactions on Search Engine Result Page~(SERP)) to some degree.

What we add on top of existing work is that we explore whether brain signals can guide the RF process and improve document re-ranking. 
\textcolor{blue}{
Recently, BCI applications have served as a new medium for human-system interaction in various real-life information systems, such as education~\cite{desoto2023utilization}, entertainment~\cite{de2023research}, virtual reality~\cite{tauscher2019immersive}. 
Therefore, understanding how BCI can enhance search performance, as it represents a typical channel for information acquisition, holds significant importance, particularly when considering research that indicates a BCI-enhanced search system is becoming realistic~\cite{11chen2022web}. 
Our findings shed light on the specific scenarios and mechanisms through which BCI can elevate the search experience via RF.
}

%% file: 3_data_collection.tex
\section{Data Collection}
\label{3}
To verify the effectiveness of brain signals and the proposed RF framework, we conducted a user study that simulates a practical search process. 
The participants were required to perform search tasks by interacting with Web pages, and their brain signals were collected during this process. 
The user study adheres to the ethical procedures approved by the ethics committee of the School of Psychology at Tsinghua University. 
In compliance with established ethical guidelines, we have taken multiple measures to safeguard the privacy of the participants, including anonymizing the collected data, obtaining informed consent from participants before the study, and allowing the participants to interrupt the experiment at any time.

\subsection{Participants}
\label{3.1}

We recruited 21 participants~(8 females and 13 males) aged from 19 to 27~(M~\footnote{Mean value.\label{M}} = 23.85, SD~\footnote{Standard deviation.\label{SD}} =2.28) from a Chinese public university. 
\textcolor{blue}{The amount of participants is analogous to prior EEG-based studies~(e.g., 18 in \cite{72ye2022don} and 23 in \cite{56pinkosova2020cortical}) and the estimated
sample size for the feature analysis in Section~\ref{5.1} is 18~(statistical
power=0.8, $\alpha$=0.05).} 
All participants are native Chinese speakers and self-reported that they are familiar with the usage of search engines. 
\textcolor{blue}{The experimental procedures encompass a one-hour neurological experiment, resting periods  totaling thirty minutes, a thirty-minute session for hair shampooing~(before and after the main experiment), a thirty-minute duration for questionnaires, experimental instructions, and wearing the EEG cap.}
The participants were compensated 15.0\$ per hour, amounting to a total payment of approximately 37.5-45.0 dollars.
 
\input{meta/Table_1.tex}

\subsection{Stimuli Preparation} 
\label{Stimuli Prepartion}
For the lab study, we constructed a dataset containing 100 queries and their corresponding documents~(an average of 39.5 documents for each query). 
The dataset is publicly available~\footnote{https://github.com/THUIR/Brain-Relevance-Feedback~\label{github}}.

\textbf{Query and document set construction.} 
We selected queries from NTCIR-11 IMine~\cite{43liu2014overview} and TREC-2009 Million Query Track~\cite{14clarke2009overview}. 
We chose these datasets for the following reasons: 
(1)~The datasets are independent of one another. 
(2)~Most of their queries are short and on a broad topic, which means that the potential benefits of RF could be easily observable. 
100 candidate queries are selected from these datasets, 50 from NTCIR-11 IMine and 50 from TREC-2009 Million Query Track. 
For NTCIR-11 IMine, all Chinese queries~(a total of 50) are included. For TREC-2009 Million Query Track, we selected 50 queries with familiar topics among Chinese users~(e.g., ``tea'', and ``pen''), and translated them into Chinese. 
However, we did not include queries with specific entities that may not be familiar to all Chinese users, e.g., ``Kansas City'', ``Vonage'', and ``Chicago Defender''.
\textcolor{blue}{
Table~\ref{tab:examples} displays sample queries, while the complete query dataset is publicly accessible\textsuperscript{\ref{github}}.
}

After that, we utilized a popular Chinese search engine Sogou~\footnote{www.sogou.com.} to retrieve the corresponding documents for each query. 
Note that we can not reuse Sogou’s datasets~(e.g., SRR~\cite{76zhang2018relevance}) since they only contain plain text of documents. 
\textcolor{blue}{This limitation extends to the absence of images in landing pages and SERP, which are required to construct the user study system. 
To address this gap, we crawl the top five SERPs by submitting these queries to Sogou.}
Then, for each document, we extracted its snippet~(document abstraction in the SERP) and crawled its landing page~(the Web page that a user is directed to after clicking the document). 
The crawling process proceeded with the Python package ``selenium''. 
All web pages were manually checked, and we dropped documents whose landing pages were invalid or failed in crawling. 
This process generates an average of 39.5 documents for each query, and each document consists of snippet content and landing page content.

\textbf{Task description construction.} 
As most of the queries are short and broad, we generated 2-5 possible task descriptions for different subtopics regarding the query, as shown in Table~\ref{tab:examples}. 
For example, for the query ``prophet'', the task descriptions include: (i)~find information about ``prophet'' in Islam, (ii)~explore the concepts of ``prophet'' in the general domain, and (iii)~search for a movie named ``prophet'' with high-quality audio source. 
To generate the task descriptions, we adopted a similar procedure as \citet{43liu2014overview} proposed in the subtopic mining subtask. 
Two Ph.D. students majoring in information retrieval are recruited as domain experts to generate the subtopics for each query with the following steps: first, they independently clustered the documents into at most five groups and generated a task description for each group according to the common subtopic of the documents. Second, they were required to compare their clustering documents and the corresponding task descriptions. After the comparison, similar task descriptions were preserved and merged, while some dissident task descriptions were discarded. Finally, they discussed and modified each task description to reach an agreement. As a result, the construction process averagely generated 2.6~(SD\textsuperscript{\ref{SD}}=1.0) task descriptions for each query.

\textbf{Snippet Annotation.} To evaluate the RF performance, we recruited 9 additional annotators with experience in search engine usage from a Chinese public university. They are required to judge the relevance of each document’s snippet regarding the task descriptions given by the experts. The judgment is binary, i.e., relevant or irrelevant. Each snippet was judged by 3 annotators, and the majority vote decided whether it was relevant. 
The Fleiss’s $\kappa$ of relevance judgment from the annotators is 0.76, reaching a substantial agreement. Note that we did not judge the relevance of the landing pages in advance due to the prohibitive cost. Instead, the participants were required to judge the relevance of their examined landing pages after the search process.

\subsection{Experimental Procedure}
\textbf{Overall pipeline.} 
The user study began with a questionnaire to collect demographic information~(e.g., age, major) and search habits~(e.g., how often do you use search engines?). 
Then, participants were instructed on the main task procedure and were informed that they could terminate the experiment and receive payment according to the time duration~(13\$ per hour). 
Next, participants were required to finish a training process, which included two search tasks to get familiar with the search process. 
After that, the main task began, in which the participants were supposed to accomplish as many search tasks as possible in 1.5 hours. 
During the main task, they were allowed to rest after every 5 tasks. 
Averagely, one participant will accomplish 36.8~(SD\textsuperscript{\ref{SD}}=10.3) search tasks in one hour. 
Finally, the participants need to fill in a post-questionnaire about their search experience.

\begin{figure*}[t]
  \centering
  \includegraphics[width=1\linewidth]{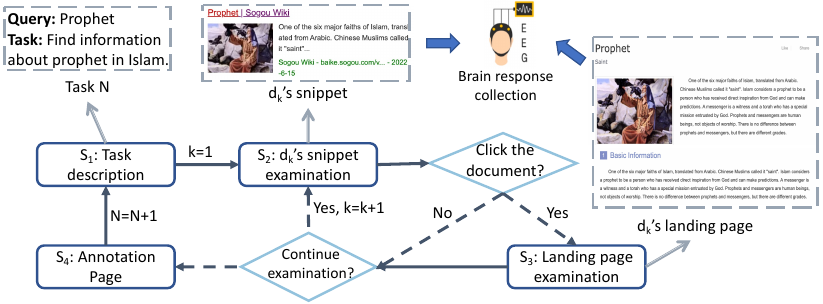}
  \caption{The structure of the main task exemplified by the query ``Prophet''~(iMine 0001).~(zoom in for details of example snippet and landing page).\label{fig:user_study}} 
\end{figure*}

\textbf{Main task procedure.} 
\textcolor{blue}{Figure~\ref{fig:user_study} delineates the sequential procedures, spanning $S_1$ to $S_4$, of a main task. 
The interface of the user study system is expounded in Section~\ref{Interface of the lab study} and visually represented in Figure~\ref{fig:interface}. 
The system can be deployed using our publicly available code\textsuperscript{\ref{github}}.}

($S_1$)~The participants view a query and one of its corresponding task descriptions. 
To mitigate the incidental effect related to the search task, we randomized the selection process of the queries, the task descriptions, and the displaying sequence of documents.
\textcolor{blue}{Examples of task descriptions are shown in Table~\ref{tab:examples}.}
 
($S_2$)~A fixation cross is presented for 0.5 seconds to indicate the location of the forthcoming stimulus. 
Then a document’s snippet extracted from the SERP~(a snippet includes a title, an abstract, and an optional image, as shown in Figure~\ref{fig:user_study}) is shown to the participant. 
The document’s snippet is set unclickable in the first 2 seconds, and then it will be wrapped by a highlighted border, indicating that it becomes clickable. 
This procedure, following existing works~\cite{51moshfeghi2016understanding,72ye2022don}, ensures that brain activity related to the motor response of clicking would not be contained during the first 2 seconds. 
\textcolor{blue}{The screenshots of the snippet page are presented in Figure~\ref{fig:serp0} and Figure~\ref{fig:serp2}, respectively.}
Note that there is some strictness in this step to ensure the presented information aligns with the participant’s perception. Similar restrictions also apply to previous studies~\cite{5azzopardi2021cognitive,18davis2021collaborative,23eugster2014predicting}. 
If the participant clicks the document, they will enter the document’s landing page~(jump to $S_3$). 
Otherwise, they can continue the examination~(stay in $S_2$ with the next document’s snippet presented) or end the search process~(jump to $S_4$).
\textcolor{blue}{To be consistent with the realistic search process, we allow the participants to revisit or click prior documents at any time in $S_2$. 
However, we only collect brain responses to each document during the first examination for further analysis.}

($S_3$)~The landing page crawled from the corresponding document’s link is presented, which generally contains more specific information than the snippet on the SERP~(as shown in Figure~\ref{fig:user_study}). 
Similar to $S_2$, the examination will last for at least 2 seconds, and the participant can choose to continue the examination~(jump to $S_2$ with the next document’s snippet presented) or end the search~(jump to $S_4$).

($S_4$)~The documents examined by the participant will reappear to collect the participant’s relevance judgment. 
The judgment is conducted with a four-point Likert scale~(ranging from 1~(``totally irrelevant'') to 4~(``perfectly relevant''). 
\textcolor{blue}{The participants are required to annotate the relevance of the landing pages~(in $S_3$) and the snippets~(in $S_4$) independently since sometimes the relevance of a document's snippet may not be aligned with its landing page.
For instance, there may be occasions where the snippet appears enticing, yet the corresponding landing page fails to meet the user's actual needs~\cite{45lu2018between}.}
For example, when the snippet relevance is 4, the landing page relevance of the same document can be judged as 1 if the landing page does not satisfy the participant~(i.e., ``bad click'' happens). 

\input{meta/interface}

 \subsection{Apparatus \& EEG Preprocessing} 
 \label{Preprocessing}
The user study proceeded with a desktop computer~(monitor size: 27-inch, resolution: 2,560×1,440) and the Google Chrome browser. 
A Scan NuAmps Express system~(Compumedics Ltd., VIC, Australia) and a 64-channel Quik-Cap~(Compumedical NeuroScan) are deployed based on the International 10–20 system to capture the participant’s EEG data~(electrical activity of the brain). 
The impedance of the channels was calibrated under 25 $k\Omega$  and the sampling rate was set at 1,000 Hz.

To standardize the preprocessing process, we segmented 2,000 ms of EEG data upon the stimulus~(i.e., snippet or landing page) presented to the participant. 
After that, we processed the EEG data with the following steps: re-referencing to averaged mastoids, baseline correlation with pre-stimulus periods~(500ms), low-pass of 50 Hz and high-pass of 0.5 Hz filtering, and down-sampling to 500 Hz. 
The preprocessing process is analogous to prior literature~\cite{56pinkosova2020cortical,72ye2022don} excepting that time-wasting artifacts removal techniques such as parametric noise covariance model~\cite{27huizenga2002spatiotemporal} and Independent Component Analysis~(ICA)~\cite{64winkler2011automatic} were not adopted. 
This is because we aim to assess the possibility of building a real-time RF method with brain signals while maintaining acceptable preprocessing time.
 
\subsection{Questionnaire analyses}
 We investigate the participant’s user study experience with a post-questionnaire. 
 We ask the participants to report their perceptual difference between the user study procedure and their daily search procedure using a five-level Likert scale~(very small, small, neither small nor big, big, and very big). 
 Most participants~(57.1\%) feel the difference is small, followed by neither small nor big~(23.8\%), big~(14.3\%), and very small~(4.8\%). 
 This indicates that our user study design is close to reality. 
 We also asked the participants about the major differences they felt, of which not allowing query reformulation is reported most often. As the first step to explore relevance feedback with brain signals, we limited the user study to an ad-hoc search scenario and left the study of multi-turn retrieval as future work.

\begin{figure*}[t]
  \centering
  \includegraphics[width=0.7\linewidth]{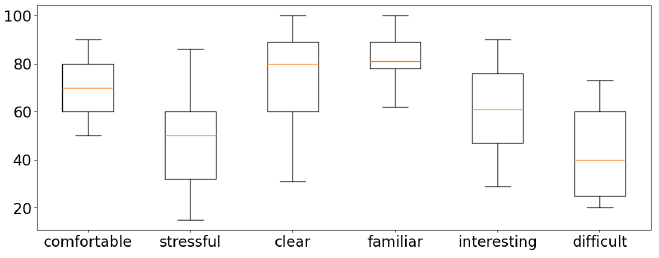}
  \caption{Box plot of participants’ evaluations on their search experience. The orange diamond represents the mean value.\label{fig:boxplot}} 
\end{figure*}

Besides, we collect participants' evaluations on their search experience by asking ``The search tasks in the user study is [comfortable/stressful/clear/familiar/interesting/difficult]'' with a 100-grade scale where 100 indicates ``strongly agree'' and 0 indicates ``strongly disagree''. 
Figure~\ref{fig:boxplot} presents the box plot of the participants’ evaluation gathered from the questionnaires. These results indicate that participants found the tasks comfortable, clear, familiar, and interesting but neither stressful nor difficult.

\subsection{Statistics of the collected data}

The dataset consists of 979 search tasks collected from 21 participants. 
The participants averagely accomplished 46.6~(SD\ref{SD}=16.6) search tasks. 
Within a search task, they averagely examined 10.9 documents~(10,670 documents in total) and clicked 1.9 of them~(1,820 clicks in total). 
According to the participants’ annotations, the average relevance of the document’s snippets is 1.72, while that of landing pages is 2.67~(note that only clicked documents get landing page annotations). 
The annotations made by the participants for the 10,670 document snippets show a correlation with the third-party annotation at Kendall’s $\tau$ of 0.66~($p$ \textless $1e^{-3}$). 
This demonstrates that there is a high level of consistency between participants and third-party annotators in their judgment of the relevance of a document. 
Besides, the datasets contain the participants’ brain responses, \textcolor{blue}{amounting to 10,670 and 1,820 EEG segments collected during their examination on snippets and landing pages, respectively.}

%% file: meta/Table_1.tex
\begin{table}
\caption{Examples of the user study queries~(translated into English).\label{tab:examples}}
\setlength{\tabcolsep}{3mm}{
\begin{tabular}{@{}p{2cm}!{\color{lightgray}\vrule}c!{\color{lightgray}\vrule}p{8.5cm}}
\specialrule{0em}{1pt}{1pt}
\toprule
Query & Source&  Task descriptions  \\ \midrule 
prophet & iMine 0001 &  (i)~find information about ``prophet'' in Islam; (ii)~explore the concepts of ``prophet'' in the general domain; (iii)~search for a movie named ``prophet'' with high-quality audio source. \\ \midrule
Persian cat & iMine 0002 & (i)~learn about the concept and characteristics of Persian cat;(ii)~download pictures of Persian cat; (iii)~learn about the market price of Persian cat;(iV)~read books with Persian cat themes. \\ \midrule
multiplication tables & iMine 0042 &(i)~download multiplication tables; (ii)~learn the tips for memorizing multiplication tables. \\ \midrule
pen & TREC 20582 & (i)~learn about fountain pens and their origins; (ii)~check the brand and price of fountain pen. \\ \midrule
tea & TREC 20906 & (i)~learn about the benefits of tea, such as nutritional components and efficacy; (ii)~learn about the varieties of tea for a course presentation; (iii)~explore the preparation methods of tea; (iv)~exploring different commercial brands of tea. \\ \midrule
milk experiments & TREC 21092 & (i)~learn information related to chemical reactions between litmus and milk; (ii)~learn information related to sensory experiment on milk or yogurt.; (iii)~learn about general science experiments related to milk. \\ \bottomrule
\end{tabular}
}
\end{table}

%% file: meta/interface.tex
\begin{figure}[h]
  \centering
  \setlength{\tabcolsep}{30pt}
  \begin{subfigure}{0.47\textwidth}
    \includegraphics[width=\linewidth]{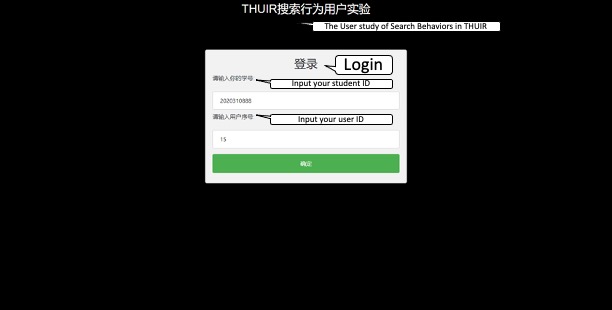}
    \caption{The login page.\label{fig:login}}
  \end{subfigure}
  \hspace{0.3cm}
  \begin{subfigure}{0.47\textwidth}
    \includegraphics[width=\linewidth]{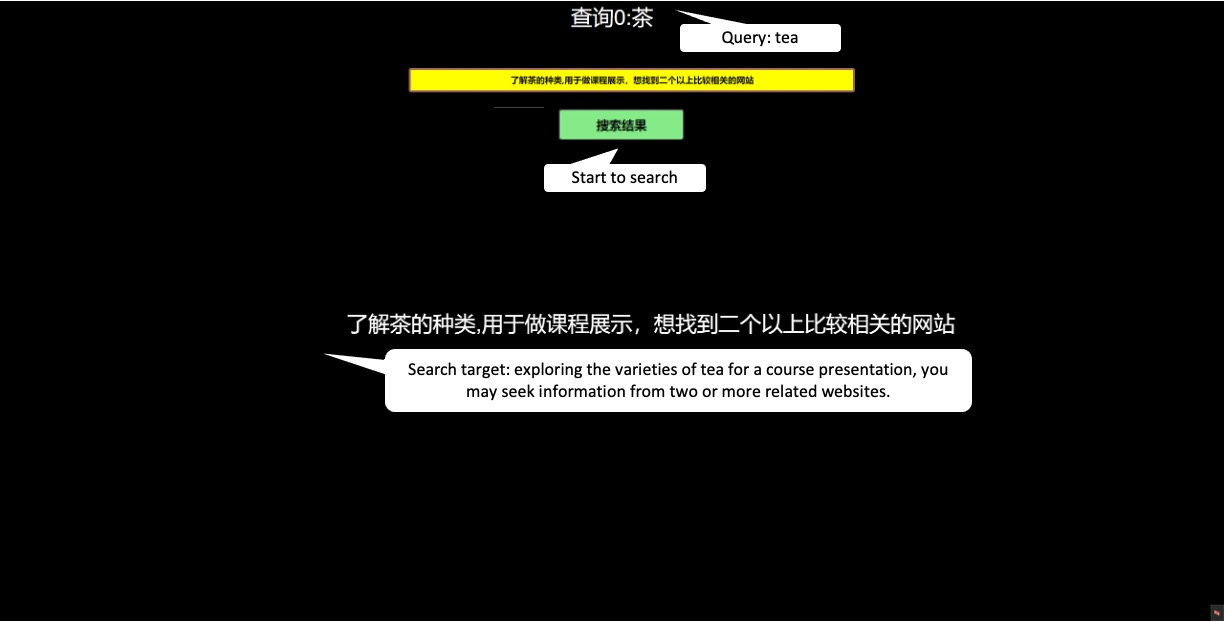}
    \caption{$S_1$: The search task description page.\label{fig:search_task}}
  \end{subfigure}\\
  \begin{subfigure}{0.47\textwidth}
    \includegraphics[width=\linewidth]{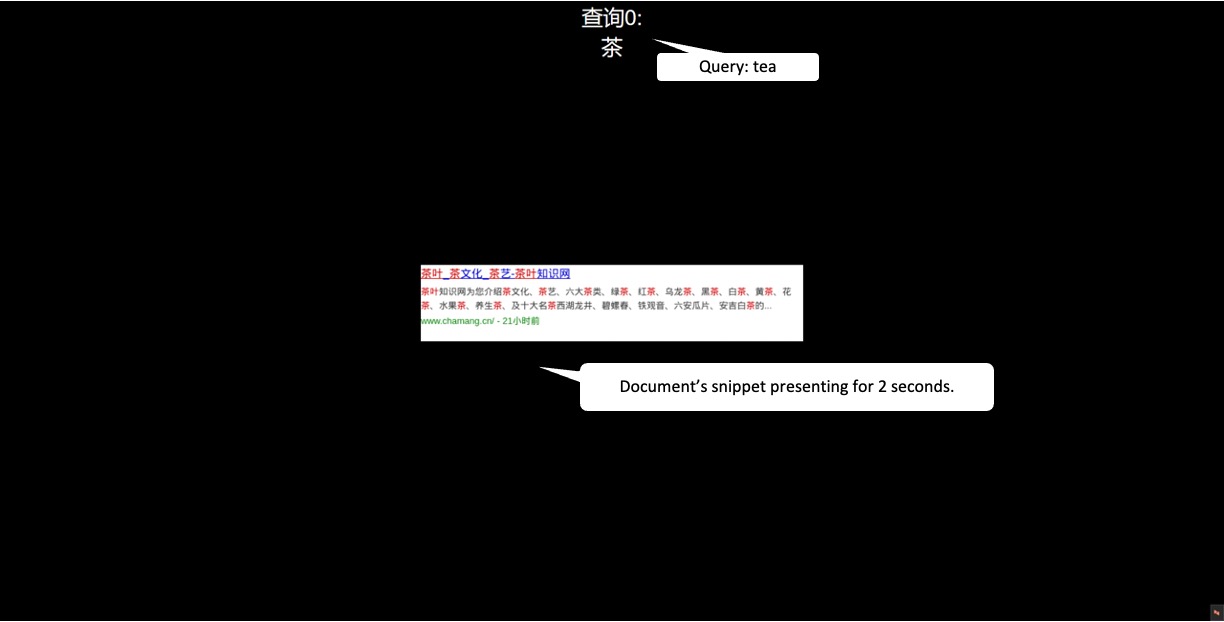}
    \caption{$S_2$: The snippet page~(within first 2 seconds)~.\label{fig:serp0}}
  \end{subfigure}
  \hspace{0.3cm}
  \begin{subfigure}{0.47\textwidth}
    \includegraphics[width=\linewidth]{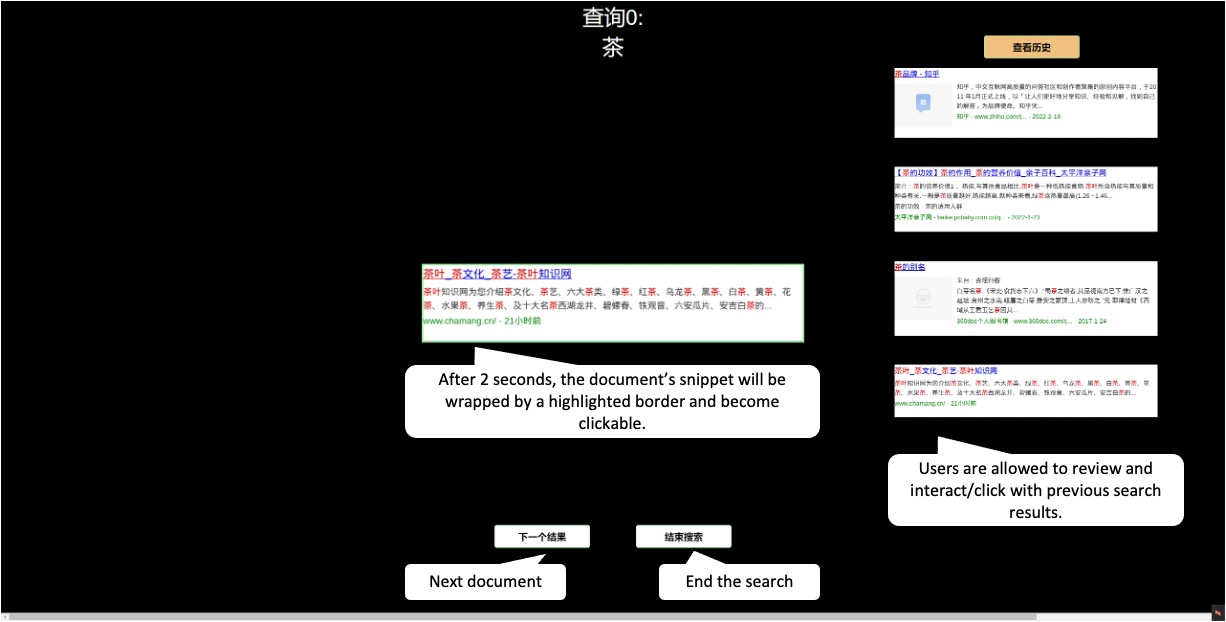}
    \caption{$S_2$: The snippet page~(after 2 seconds)~.\label{fig:serp2}}
  \end{subfigure}\\
  \begin{subfigure}{0.47\textwidth}
    \includegraphics[width=\linewidth]{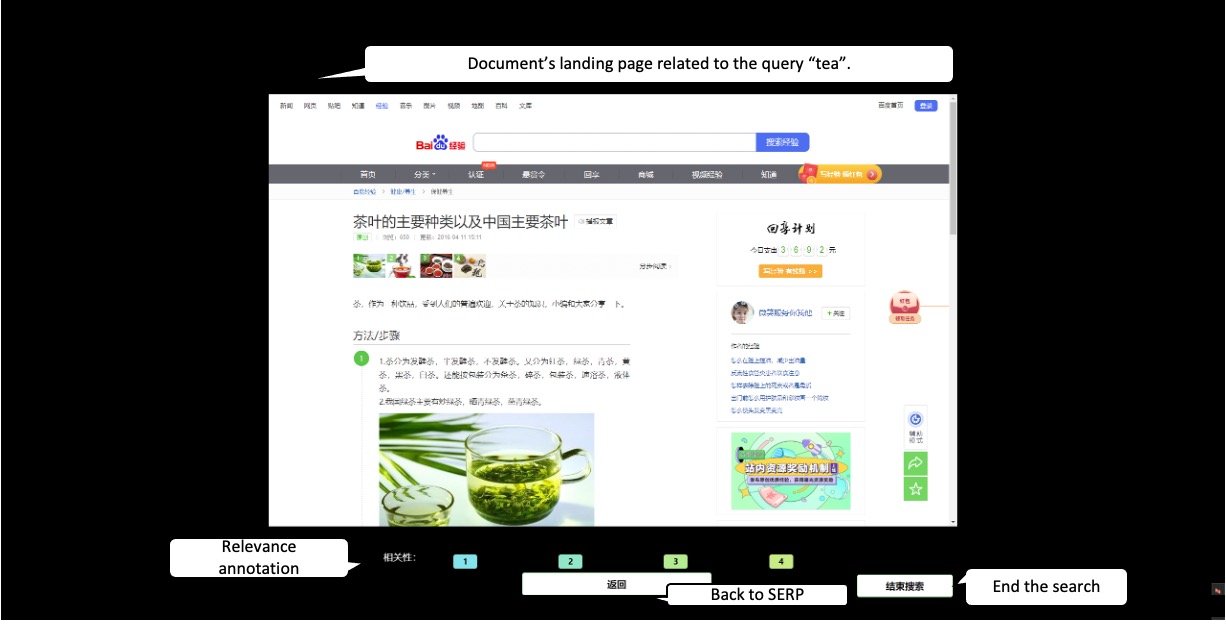}
    \caption{$S_3$: The landing page.\label{fig:land}}
  \end{subfigure}
  \hspace{0.3cm}
  \begin{subfigure}{0.47\textwidth}
    \includegraphics[width=\linewidth]{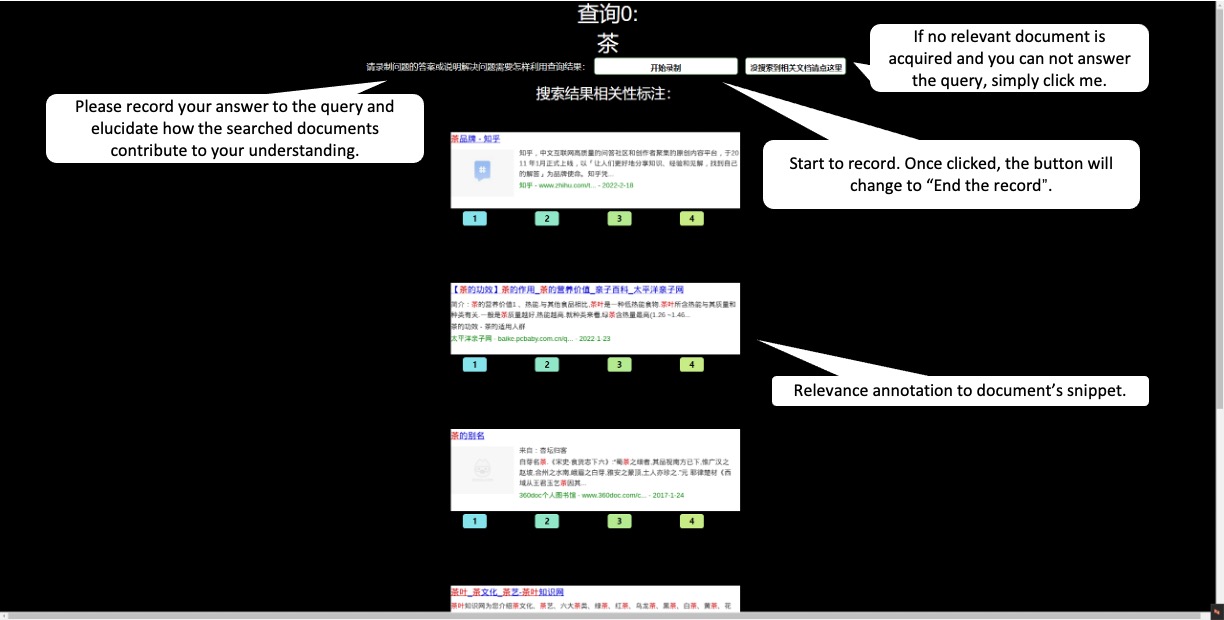}
    \caption{$S_4$: The annotation page.\label{fig:anno}}
  \end{subfigure}
  \caption{\textcolor{blue}{Interface of the lab study. Figure~\ref{fig:login} depicts the login page. Figure~\ref{fig:search_task} illustrates the search task description page corresponding to step $S_1$ in the main task~(detailed in Section~\ref{3}). 
  Meanwhile, Figures~\ref{fig:serp0} and~\ref{fig:serp2} correspond to step $S_2$, Figure~\ref{fig:land} corresponds to step $S_3$, and Figure~\ref{fig:anno} corresponds to step $S_4$.\label{fig:interface}}}
\end{figure}

\subsection{\textcolor{blue}{Interface of the lab study}}
\label{Interface of the lab study}

\textcolor{blue}{
Figure~\ref{fig:interface} showcases the interface of the Web search lab study using an example query ``tea''~(TREC 20582). On all pages, the background color is set to black, a common practice in neurological experiments to minimize interference from irrelevant factors~\cite{mudrik2010erp}.
}

\textcolor{blue}{
In Figure~\ref{fig:login}, the login page is displayed, where participants enter their student ID~(or a randomly generated ID if they are not students) along with a randomly assigned user ID. 
Post login, search tasks are sequentially presented to the participants. 
To complete each search task, participants undertake four steps, elaborated in Section~\ref{3} and Figure~\ref{fig:user_study}.
}

\textcolor{blue}{
Figures~\ref{fig:search_task}-\ref{fig:anno} provide illustrations and screenshots for steps $S_1$-$S_4$~(as detailed in Section~\ref{3}). 
Figure~\ref{fig:search_task} portrays the search task description page where participants are instructed to thoroughly read the query term~(e.g., ``tea'') at the top, followed by the task descriptions~(e.g., ``exploring the varieties of tea for a course presentation, seeking information from two or more related websites''). Subsequently, participants can click ``start to search'' to proceed to step $S_2$. 
Figures~\ref{fig:serp0} and~\ref{fig:serp2} depict the document snippets page in step $S_3$. 
Initially, for the first two seconds, the document’s snippet is rendered unclickable~(as demonstrated in Figure~\ref{fig:serp0}) to ensure the exclusion of brain activity related to the motor response of clicking~\cite{51moshfeghi2016understanding,72ye2022don}. 
Post this period, participants are provided with four optional actions, illustrated in Figure~\ref{fig:serp2}: (1) click the document to access the corresponding landing page, (2) click ``Next document'' to review the following document, (3) click ``End the search'' to proceed to the annotation page, (4) examine and interact (e.g., click) with previous documents displayed on the right side of the screen. 
Figure~\ref{fig:land} demonstrates the landing page interface where participants can scroll to inspect all content on the landing page. 
Upon examination, participants may annotate a relevance score for the landing page and return to $S_2$ by clicking ``Back to SERP''. 
If participants end the search in $S_2$ or $S_3$, they are directed to the annotation page as illustrated in Figure~\ref{fig:anno}. 
On this page, participants are supposed to verbalize a concise answer to the search tasks, following which they are required to annotate each document snippet with a relevance score.
}

%% file: 4_method.tex
\section{method}
\label{4}
\begin{figure*}[t]
  \centering
  \includegraphics[width=0.95\linewidth]{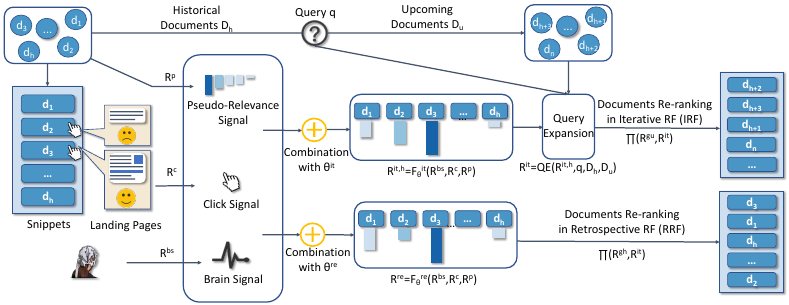}
  \caption{
  During the search process, the user’s brain signals and click signals are collected in addition to the pseudo-relevance signals provided by the system. 
  These signals are then combined into two relevance scores for IRF and RRF tasks, respectively. 
  Then the combined relevance score for RRF is directly applied to re-estimating the relevance of the historical documents. 
  Besides, the upcoming documents can be re-ranked with a query expansion module which extracts useful contents in historical documents with the combined relevance score for IRF.\label{fig:model}} 
\end{figure*}

\textcolor{blue}{This section elaborates the RF problem definition, the steps of the proposed RF framework~(as visualized in Figure~\ref{fig:model}), and the training and evaluation pipeline of RF.}

\input{meta/Table_notations.tex}

\subsection{Problem definition}
\label{4.1}
We assume a user issues a query $q$, and the documents list associated with the query $q$ is $\mathcal{D}=\{d_1,d_2,...d_n\}$. 
Suppose the user examined $h_{max}$ documents under the query $q$ before they end the search. 
At a certain state during the process, $h$~(from 1 to $h_{max}$) historical documents $\mathcal{D}_h=\{d_1,d_2,...,d_h\}$ have already been examined, and there still exist $n-h$ up-coming documents $\mathcal{D}_u=\{d_{h+1},d_{h+2},...,d_n\}$ have not been examined.

With the feedback signals acquired from the interactions on historical documents $\mathcal{D}_h$, the proposed framework aims to facilitate two RF tasks, i.e., iterative RF~(IRF) and retrospective RF~(RRF).
IRF re-ranks upcoming documents $\mathcal{D}_u$ by re-estimating their relevance $R^{it} =\{r^{it}_{h+1} ,...,r^{it}_{n}\}$.
It can be adopted in real-time search scenarios: as the number of examined documents $h$ increases, IRF iteratively re-ranks $\mathcal{D}_u$ to benefit the current search process. On the other hand, RRF re-ranks historical documents $\mathcal{D}_h$ by re-estimating their relevance $R^{re} = \{r^{re}_{1} ,...,r^{re}_{h}\}$. It does not aim at re-ranking documents for the current search process, but it can be utilized to provide a better ranking list in prospective search processes with similar intents.

To evaluate IRF and RRF, we apply ranking-based metric $\Pi$~(e.g., Normalized Discounted Cumulative Gain~(NDCG)~\cite{29jarvelin2002cumulated}). 
Suppose the ground truth relevance for $\mathcal{D}_u$ and $\mathcal{D}_h$ are $R^{gu}$ = $\{r^{gu}_{h+1}, ..., r^{gu}_n \}$ and $R^{gh}$ = $\{r^{gh}_1, ..., r^{gh}_h \}$, respectively, where $r^{gu}_i$~($i > h$) or $r^{gh}_i$ ($i \leq h$) is the ground truth relevance of the $i^{th}$ document. 
The performance of IRF and RRF can be measured as $\Pi({R^{gu},R^{it}})$ and $\Pi({R^{gh},R^{re}})$, respectively.

\subsection{RF Framework}
The proposed RF framework includes the following steps: 
(1)~Pseudo-relevance signals, click signals and brain signals are independently transformed into base relevance scores $R_p$, $R_c$, and $R_{bs}$, respectively. 
The brain-based relevance score $R_{bs}$ is chosen from the brain-based relevance scores of the document’s snippet $R^s$ and the document’s landing page $R^l$. 
\textcolor{blue}{
(2)~For the $i^{th}$ historical document $d_i \in \mathcal{D}_h$, the feedback signals $r_i^p\in R_{p}$, $r_i^c\in R_{c}$, and $r_i^{bs} \in R_{bs}$, are combined into relevance scores $r^{it,h}_i \in R^{it,h}$ and $r^{re}_i\in R^{re}$ for IRF and RRF, respectively. 
}
For simplicity, we utilize $F_\theta$ to denote the combination using parameter $\theta$, e.g., $R^{it}$ = $F_{\theta^{it}}(R_{bs}, R_c, R_p)$. 
(3)~In RRF, the historical documents $\mathcal{D}_h)$ are directly re-ranked according to the combined relevance score $R^{re}$. 
(4)~In IRF, we generate the relevance score for upcoming documents by a query expansion module $QE$, which can be formulated as $R^{it}$ = $QE(R^{it,h}, q, \mathcal{D}_h, \mathcal{D}_u)$, where the combined relevance score for historical documents $R^{it,h}$ are applied to balance the importance of each document for query expansion module $QE$.

\subsubsection{Signals preparation}
To generate high-quality RF, we acquire and combine base relevance scores independently estimated from pseudo-relevance signals, click signals, and brain signals, which are detailed as follows:

\textbf{Pseudo-relevance score.}
By measuring the semantic similarity between $q$ and each document $d_i \in \mathcal{D}_h$, we generate the pseudo-relevance scores $R_p=\{r^p_1, r^p_2,...,r^p_h\}$, where $r^p_i \in [0,1]$ is the ranking score of $d_i$. 
For a query $q$ and a document $d$, we adopt a BERT re-ranker~\cite{41yates2021pretrained,53nogueira2019passage} to measure their semantic similarity, denoted as $BERT(q,d)$. 
The BERT re-ranker is initialized by fine-tuning BERT-Chinese~\footnote{https://github.com/ymcui/Chinese-BERT-wwm} on $T^2Ranking$ dataset~\footnote{https://github.com/THUIR/T2Ranking} with the same procedures and available codes in the dataset’s original paper~\cite{66xie2023t2ranking}. 
\textcolor{blue}{
The dataset contains human annotations on queries and documents extracted from Sogou’s search log, which is similar to the dataset constructed for our user study. 
Besides, we use the snippet content to represent the document 
$d$'s for estimating the semantic similarity, as it reflects what users typically observe on SERPs.
This choice effectively captures the primary theme of the landing page while filtering out extraneous or irrelevant details.
}

\textbf{Click-based relevance score.}
The click-based relevance score $R^c=\{r^c_1,r^c_2,\ldots,r^c_h\}$ is generated from user's clicks behaviors, where $r^c_i=0$~(or $1$) indicates the user abandons~(or clicks) the $i^{th}$ document.

\textbf{Brain-based relevance score.} \label{Brain-based relevance score}
With EEG devices, the user's brain responses to snippet content and landing page content are collected as $\mathcal{X}^s=\{x^s_1,x^s_2,\ldots,x^s_h\}$ and $\mathcal{X}^l=\{x^l_1,x^l_2,\ldots,x^l_h\}$, respectively, where $x^s_i$ and $x^l_i$ are the brain responses to the $i^{th}$ document's snippet and landing page, respectively. 
\textcolor{blue}{A sample of the brain signals, denoted as $x \in \{x^s_i,x^l_i\}$, can be presented as a vector in  the space $\mathbb{R}^t$, with $t$ denoting the length of EEG features.} 
With a brain decoding model $G$~(elaborated in Section~\ref{5.1}), $X^s$ and $X^l$ are then transformed into brain-based relevance $R^s=\{r^s_1,r^s_2,\ldots,r^s_h\}$ and $R^l=\{r^l_1,r^l_2,\ldots,r^l_h\}$, respectively, where $r^s_i$ and $r^l_i$ are real numbers in $[0, 1]$ indicating the brain-based relevance score for the snippet and landing page, respectively. 
As the user's brain responses to the landing page content, represented by $x^l_i$, cannot be acquired when they do not click the document and enter the landing page, the corresponding value of $r^l_i$ is masked and considered as unavailable in such situations.

Based on $R^l$ and $R^s$, we generate $R^{bs}=\{r^{bs}_1,\ldots,r^{bs}_h\}$, which presents the brain-based relevance scores of the documents $\mathcal{D}_h$. We use different principles to generate $R^{bs}$ for IRF and RRF, respectively. 
The reasons and details are as follows:

\begin{enumerate}
  \item[(i)] In IRF, we extract information from $\mathcal{D}_h$ in the query expansion procedure. 
  As the snippet usually presents the document more briefly and contains less noise than the landing page, we utilize the snippet content rather than the landing page content to represent the document. Hence, we can simply assign $r^{bs}_i = r^s_i$.
  
  \item[(ii)] In RRF, we aim to re-rank the documents $\mathcal{D}_h$ themselves. If a document $d_i$ has an attractive snippet but a landing page of low quality, it often leads to a poor user experience~\cite{45lu2018between} and should not be evaluated as a satisfying document. 
  Therefore, if $r^l_i$ is available~(not a masked value), $r^{bs}_i$ is assigned the value of $r^l_i$; otherwise, we assign $r^{bs}_i$ as $r^s_i$.
\end{enumerate}

\subsubsection{Relevance score combination}

For the $i$-th document, its combined relevance scores $r^{it}_i$ and $r^{re}_i$ are combined with various relevance scores from pseudo-relevance signals~($r^p_i$), click signals~($r^c_i$), and brain signals~($r^{bs}_i$), which can be formulated as:

\begin{equation}
	r_{i}^{it,h}  =  \theta ^ {it,bs} \cdot r_{i}^{bs} + \theta^{it,c} \cdot   r_{i}^{c} + \theta^{it,p}   \cdot  r_ {i}^ {p}
\end{equation}

\begin{equation}
 r_{i}^ {re}  =  \theta ^ {re,bs}  \cdot   r_{i}^ {bs}  +  \theta ^ {re,c}  \cdot  r_{i}^ {c}  +  \theta ^ {re,p}  \cdot   r_{i}^ {t} 
\end{equation}

Where $\Theta^{it} = \{\theta^{it,bs}, \theta^{it,c}, \theta^{it,p}\}$ and $\Theta^{re} = \{\theta^{re,bs}, \theta^{re,c}, \theta^{re,p}\}$ are the combination parameters. It's critical to understand that we use different combination parameters for IRF and RRF for the following reasons:

\begin{enumerate}
  \item[(i)] We utilize different principles to generate $R^{bs}$ for IRF and $R^{bs}$ for RRF, hence the combination weights should be different.
  
  \item[(ii)] $R^{it,h}$ is applied as an indication to extract text information from $\mathcal{D}_h$'s snippet while $R^{re}$ is directly applied to re-estimate the relevance of $\mathcal{D}_h$. 
  Hence inherent differences exist between them, e.g., $R^{it,h}$ emphasizes whether the document's snippet content is relevant while $R^{re}$ should also reflect the document's quality beyond the text.
\end{enumerate}

Besides, it is also important to note that $R^{it,h}$ and $R^{re}$ can be either fixed or adjustable to different search scenarios, which are detailed in Section~\ref{5.2.1} and Section~\ref{5.4}, respectively.

\begin{figure*}[t]
  \centering
  \includegraphics[width=0.88\linewidth]{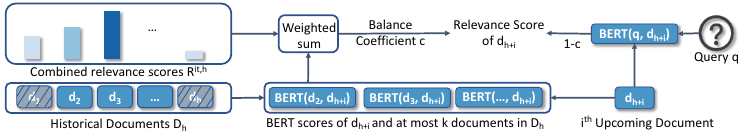}
  \caption{The query expansion module.\label{fig:qe}} 
\end{figure*}

\subsubsection{Query Expansion}

This section elaborates on the query expansion module $QE$, \textcolor{blue}{as depicted in Figure~\ref{fig:qe}}.
The query expansion module utilizes the combined relevance score $R^{it,h}$ and the historical documents $\mathcal{D}_h$ to re-rank upcoming documents $\mathcal{D}_u$. 
As BERT has become a prevalent base ranker for RF~\cite{42lin2021multi,74yu2021improving,78zheng2020bert}, we adopt a BERT-based query expansion module into our RF framework with the following extensions:
(i)~We extend this module into an interactive setting for IRF.
(ii)~We bring in the estimated relevance score combined from various signals.
The details of the query expansion module are as follows:

\textcolor{blue}{First, we select a maximum of $k$~(a fixed value set as 10) documents from the historical documents $\mathcal{D}_h$, denoted as $\mathcal{D}_s = \{d_{s^1}, \dots, d_{s^k}\}$, based on their combined relevance scores $r^{it,h}_i$~($i \in {1, 2, \dots, h}$), which surpass those of other documents.} 
Second, for each upcoming document $d_{h+i}$, we calculate the BERT score $BERT(d_s, d_{h+i})$ for each document $d_s$ in $\mathcal{D}_s$.
Third, the calculated BERT scores are weighted sum based on the combined relevance score $R^{it,h}$ to acquire the relevance score $r_{h+i}^f$ for the $(h + i)^{th}$ document:

\begin{equation}
r^f_{h+i} = \sum_{j=1}^k \frac{e^{r^{it,h}_{s_k}}}{\sum_{l=1}^k e^{r_{it,h}^{s_l}}}\cdot \text{BERT}(d_{s_j}, d_{h+i})
\end{equation}

Fourth, the relevance score for the $i^{th}$ upcoming document $d_{h+i}$ is trade-offed by its BERT score with the initial query $q$ and the feedback information $r_{h+i}^f$:

\begin{equation}
i_t^f = r_{h+i} \cdot c + \text{BERT}(q, d_{h+i}) \cdot (1 - c)
\end{equation}

where $c$ is a coefficient~(set as 0.1) to balance the influence of feedback documents and the initial query, which is widely applied in existing RF methods~\cite{7bi2018revisiting,58rocchio1971relevance}. 
Finally, the upcoming documents $\mathcal{D}_u$ are re-ranked with $R^{it} = \{r^{it}_{h+1}, \ldots, r^{it}_{n}\}$ to facilitate the search process.

\subsection{\textcolor{blue}{RF training \& evaluation}}
\label{5.0}

\input{meta/algorithm_2}

\textcolor{blue}{
Algorithm~\ref{algorithm:pipeline} presents the training and evaluating pipelines of the RF experiments. 
Unlike existing literature~\cite{23eugster2014predicting,72ye2022don} that splits the dataset without considering the natural sequence of data samples, we adopt a split-by-timepoint protocol to train and evaluate the RF model. 
When a new search task is presented to a participant $u$, we first prepare the brain decoding model~(shown in Algorithm~\ref{algorithm:pipeline} line 3-4) using the brain data collected from previous search tasks of the same participants~(the personalized model $G_p$) or other participants~(the global model $G_g$). 
Training a personalized model $G_p$ for each participant is necessary as brain signals vary across different individuals, as indicated by previous studies~\cite{37lan2018domain,79zhong2020eeg}. 
However, a participant's data may be insufficient for training $G_p$ at the beginning of his or her search process. 
Hence, we adopt a generalized model $G_g$ trained using other participant's data as a substitute until his or her collected data size reaches a minimum required size~(set as 100 data samples).
}

\textcolor{blue}{
After that, experiments involving IRF and RRF are executed, as delineated in lines 5-13 and 14-18 of Algorithm~\ref{algorithm:pipeline}, respectively.
For IRF, valuable RF signals are gathered from historical document $\mathcal{D}_h$, enabling the generation of relevance scores for the unseen documents $\mathcal{D}_u$ utilizing the RF method~( detailed in Section~\ref{4}).
Given the absence of user annotations for all unseen documents, IRF performance is evaluated using third-party annotations $R^{gu}$, incrementally as the number of historical document $h$ increases.
Consequently, $h_{max}$ ranking-based metrics, $\Pi(R^{gu},R^{it})$, are calculated for a query $q$.
In contrast, RRF performance evaluation is deferred until the end of the current search query due to its retrospective nature.
The re-ranking performance for $\mathcal{D}_{h_{max}}$ is calculated by $\Pi(R^{gh}, R^{re})$, where $R^{re}$ is the relevance score generated with the RF framework and $R^{gh}$ indicates the user's annotation. 
}

%% file: meta/Table_notations.tex
\begin{table}
\caption{\textcolor{blue}{Notations of the relevance feedback framework.}\label{tab:notations}}
\setlength{\tabcolsep}{3mm}{
\begin{tabular}{@{}c!{\color{lightgray}\vrule}c}
\specialrule{0em}{1pt}{1pt}
\toprule
Notation & Definition \\ \midrule
$q, \mathcal{D}$& the query, the documents regarding the query \\ \midrule
$\mathcal{D}_h, \mathcal{D}_u$ & the historical documents, the up-coming documents \\\midrule
$\mathcal{R}^p,\mathcal{R}^c$& $\mathcal{D}_h$'s pseudo-relevance scores, click-based relevance scores \\\midrule
$\mathcal{X}^s, \mathcal{X}^l,$ & brain responses to $\mathcal{D}_h$'s snippet, landing page\\\midrule 
$\mathcal{R}^s,\mathcal{R}^l$& $\mathcal{D}_h$'s brain-based relevance scores for its snippet, landing page \\ \midrule
$\mathcal{R}^{bs}$ & $\mathcal{D}_h$'s brain-based relevance scores for the document \\ \midrule
$\mathcal{R}^{it,h},\mathcal{R}^{re}$ & $\mathcal{D}_h$'s combined relevance scores in IRF, RRF \\ \midrule
$\mathcal{R}^{it}$ & $\mathcal{D}_u$'s estimated relevance scores in IRF \\ \midrule
$\theta^{it},\theta^{re}$ & combination parameters in IRF, RRF \\ \midrule
$\mathcal{R}^{gu},\mathcal{R}^{gh}$ & the ground truth relevance scores of $\mathcal{D}_u$, $\mathcal{D}_h$ \\ \bottomrule
\end{tabular}
}
\end{table}

%% file: meta/algorithm_2.tex
\begin{algorithm}[t]
\caption{\textcolor{blue}{Overall RF pipeline \& experimental setup}}\label{algorithm:pipeline}
\SetKwInput{KwInput}{Input} 
\SetKwInput{KwOutput}{Output} 
\KwInput{\textcolor{blue}{A user $u$, a series of search tasks consist of a query set $\mathcal{Q}$ and document set $\mathcal{D}$ for each query $q \in \mathcal{Q}$, a generalized brain decoding model $G_g$ trained from other participants' brain recordings, a base ranker BERT. } }

\KwData{\textcolor{blue}{A personalized brain encoding model $G_p$ share the same structure with $G_g$ but initialized with random parameters, hyper-parameters $\Theta^{it}=\{\theta^{it}_{bs}, \theta^{it}_{c},\theta^{it}_{p}\}$ and $\Theta^{re}=\{\theta^{re}_{bs}, \theta^{re}_{c},\theta^{re}_{p}\}$, all collected brain data samples $\mathcal{X_{\text{global}}}=[]$, all collected user annotations $\mathcal{R_{\text{global}}}=[]$.}}

\KwOutput{\textcolor{blue}{The averaged document re-ranking performance $S_{IRF}$ and $S_{RRF}$ for IRF and RRF, respectively. }}

\textcolor{blue}{Initialize the brain decoding model $G$ as $G_g$, and initialize both $S_{IRF}$ and $S_{RRF}$ as empty lists $[]$.}

\For{\textcolor{blue}{each $q \in \mathcal{Q}$}}{

    \textcolor{blue}{Train $G_p$ with $\mathcal{X}_{global}$ and $\mathcal{R}_{global}$.} 
    
    \textcolor{blue}{Transform $G$ to $G_p$ if the size of collected brain data samples $\mathcal{X}$ attains a total of 100 .}

    \tcp{\scriptsize Evaluating IRF performance by re-ranking unseen documents as the search proceeds.}
    
    \For{\textcolor{blue}{each $h \in \{1,...,h_{max}\}$}\tcp{\scriptsize $h_{max}$ is the number of documents user $u$ has interacted with in total.}}{ 
    
    \textcolor{blue}{Collect user $u$'s brain responses $\mathcal{X}$ corresponding to document $d \in \{d_1, ..., d_h\}$.}  
    
    \textcolor{blue}{Generate $R^{bs}$ with $\mathcal{X}$ and the brain decoding model $G$.}

    \textcolor{blue}{Generate $R^{c}$ according to user $u$'s click behaviors, calculate $R^{p}$ base on the text-based ranking scores $\text{BERT}(q,D)$.}

    \textcolor{blue}{$R^{it}=QE^{F_{\Theta^{it}}(R^{bs}, R^{c}, R^{p})}$.} \tcp{\scriptsize Generate combination relevance score for IRF.
    } 

    \textcolor{blue}{Utilize third-party relevance annotations for $d \in \{d_{h+1}, ..., d_n\}$ as $R^{gu}$.}

    \textcolor{blue}{$S_{IRF}.\text{append}(\Pi(R^{gu}, R^{it}))$. }\tcp{\scriptsize Calculate ranking-based metrics for IRF.}
    } 

    \tcp{\scriptsize Evaluating RRF performance by re-ranking historical documents after the search ends.}

    \textcolor{blue}{Collect user $u$'s brain responses $\mathcal{X}$ corresponding to document $d \in \{d_1, ..., d_{h_{max}}\}$.} 
    
    \textcolor{blue}{Generate $R^{bs}$, $R^{c}$, and $R^{p}$ for document $d \in \{d_1,...,d_{h_{max}}\}$.}  

    \textcolor{blue}{$R^{re}=F_{\Theta^{re}}(R^{bs}, R^{c}, R^{p})$.} \tcp{\scriptsize Generate combination relevance score for RRF.
    } 

    \textcolor{blue}{Collect third-party relevance annotations for $d \in \{d_{1}, ..., d_{h_{max}}\}$ as $R^{gh}$.}

    \textcolor{blue}{$S_{RRF}.\text{append}(\Pi(R^{gh}, R^{re}))$. }\tcp{\scriptsize Calculate ranking-based metrics for RRF.}

    \tcp{\scriptsize Extend $\mathcal{X}_{global}$ and $\mathcal{R}_{global}$ for the split-by-timepoint training of brain decoding model.}

    \textcolor{blue}{$\mathcal{X}_{global}.\text{append}(\mathcal{X})$, $\mathcal{R}_{global}.\text{append}(R^{gh})$.}

}

\textcolor{blue}{$S_{IRF}=Average(S_{IRF}), S_{RRF}=Average(S_{RRF})$}

\Return \textcolor{blue}{$S_{IRF}, S_{RRF}$};

\end{algorithm}

%% file: 5_experiments_results.tex
\section{EXPERIMENTS AND RESULTS}
\label{5}
We conduct empirical experiments~(all implementation codes and datasets are available\textsuperscript{\ref{github}}) to address the following research questions:

\begin{itemize}
  \item \textbf{RQ1:} Can brain signals provide guidance for RF?
  \item \textbf{RQ2:} To what extent can we improve RF with the proposed framework and brain signals?
  \item \textbf{RQ3:} How do brain signals improve RF performance in different search scenarios?
  \item \textbf{RQ4:} Can we further improve RF by adaptively adjusting the combination weight of brain signals and other RF signals?
\end{itemize}

To address \textbf{RQ1}, we explore the effectiveness of the brain decoding model, which aims to predict brain-based relevance scores $R^{bs}$. 
Then we analyze the document re-ranking performance of the proposed RF framework in IRF and RRF, respectively, to answer \textbf{RQ2}. 
Furthermore, we analyze the performance gain brought by brain signals in different search scenarios, especially in two particular cases, i.e., \textit{bad click identification} and \textit{non-click relevance estimation} to answer \textbf{RQ3}. 
We observe that the benefit brought by brain signals varies with different search scenarios. 
Therefore, the potential advancement of RF relies on designing a better combination strategy of brain signals and other RF signals. 
Hence, we then explore an adaptive method to combine RF signals depending on the search scenarios, which answers \textbf{RQ4}.

\input{5.1.tex}

\input{5.2.tex}

\input{5.3.tex}

\input{5.4.tex}

%% file: 5.1.tex
\subsection{Brain Decoding Experiment~(RQ1)}
\label{5.1}
This section elaborates on the experiments that decode brain responses to the snippet and landing page content into brain-based relevance scores. 
We first explain our experimental setup process, including feature extraction, decoding model selection, and evaluation protocols. 
Then, we show the difference between brain responses to relevant and irrelevant items with statistical analyses on the extracted features. 
Afterward, a decoding experiment is performed to evaluate the performance of relevance estimation based on brain signals.

\subsubsection{Experimental Setup}

\textbf{Feature Extraction:} We extract differential entropy~(DE)~\cite{28hyvarinen1997new} features for the brain decoding task. DE is a popular frequency-based feature in EEG-based prediction~\cite{21duan2013differential}, which has shown superior performance in emotion recognition~\cite{21duan2013differential} and satisfaction detection~\cite{71ye2022brain}. 
The DE features are calculated in five bands, i.e., $\delta$~(0.5-4 Hz), $\theta$~(4-8 Hz), $\alpha$~(8-13 Hz), $\beta$~(13-30 Hz), and $\gamma$~(30-50 Hz), on 62 EEG channels. 
As a result, a sample of EEG data is preprocessed into a vector of size $62\times5$ for each brain response upon the stimulus.

\textbf{Decoding Model Selection:} We adopt a support vector machine~(SVM) with the Gaussian kernel as the base brain decoding model since it is prevalent and effective for brain decoding~\cite{6bhardwaj2015classification,21duan2013differential}.
Besides, it requires less computing effort than neural networks~\cite{71ye2022brain,79zhong2020eeg}, and can meet the demand of online training and inferencing in realistic systems. 
Specifically, we observe that the computing time for brain decoding with an SVM is negligible compared to~(less than one-thousandth of) the BERT-based re-ranking algorithm. 
Hence, the proposed RF process could be conducted in realtime. 

\textbf{Evaluation:} 
\textcolor{blue}{
We evaluate the performance of the brain decoding model $G$~($G_p$ or $G_g$) in terms of its classification abilities for relevant and irrelevant items.
}
Since relevance annotation 1 makes up 68.2\% data samples, relevance annotation 1 is regarded as a negative sample~(irrelevant) and annotation 2-4 as positive samples~(relevant). 
Hence, the classification problem is transformed into a binary classification problem, and we measure its performance with Area Under Curve~(AUC). 
Note that we use the same classification model $G$~($G_p$ or $G_g$) for the classification of brain response to snippet and landing page since we observe that its performance is better than training independent models.

\begin{figure*}[t]
  \centering
  \includegraphics[width=0.88\linewidth]{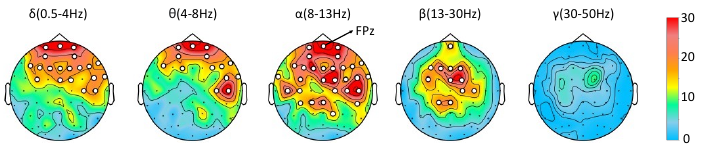}
  \caption{Topography which shows the significance of difference (F-value) between brain response to relevant/irrelevant Web pages. Highlighted channels indicate the differences are significant at $p<1 \times 10^{-3}$ level.\label{fig:topo}} 
\end{figure*}

\subsubsection{Feature Analyses}

Frequency-based EEG analyses are widely applied in relevance judgment~\cite{72ye2022don}, emotion recognition~\cite{35koelstra2011deap}, and other domains. 
We calculate the mean value for each channel obtained by averaging DE features extracted from each participant's brain responses to web pages~(both snippets and landing pages) and over two conditions: relevant and irrelevant stimuli. 
Figure 5 presents the F-value for 62 EEG channels by performing an ANOVA test, where the highlighted channels indicate the difference is significant at $p < 1 \times 10^{-3}$ level. 
We observe various significant channels in $\delta$, $\theta$, $\alpha$, and $\beta$ bands. Among these channels, the most significant difference appears in the FPz in $\alpha$ band~($F[1, 20] = 39.03$, $p = 4.2 \times 10^{-6}$, $M_{diff} = -0.40 \ln(Hz)$). 
This finding indicates that brain responses to relevant and irrelevant stimuli are distinguishable.

Additionally, we observe that in $\delta$, $\theta$, and $\alpha$ bands, the most significant differences appear in the frontal region, which is inconsistent with prior literature \cite{72ye2022don}. 
The frontal region is related to executive functions such as judgment and problem-solving~\cite{70yang2017modulating}. \citet{72ye2022don} suggests that this difference is a consequence of cognitive function and working memory executing during relevance judgment. 
On the other hand, we also observe significant neural differences in the central region, especially in the $\beta$ and $\alpha$ bands. 
\citet{3allegretti2015relevance,56pinkosova2020cortical,68yang2019late} also have similar observations in the central region. 
Despite the different settings among these studies~(stimuli based on visual~\cite{3allegretti2015relevance}, text~\cite{56pinkosova2020cortical}) and ours~(stimuli with multimedia content), the common finding in the central region indicates a potential link between brain functions in this region and the concept of relevance. 
A partial explanation of this potential link is the memory processing during relevance judgment~\cite{56pinkosova2020cortical}, e.g., recognizing whether an item is relevant by recalling knowledge from memory.

Besides these common observations, there are also several findings in contrast with prior research. 
For instance, \citet{72ye2022don} find the most significant differences in the $\beta$ band while we observe more differences in the $\alpha$ band than in the $\beta$ band. 
A likely reason for this difference is the design of the search tasks: \citet{72ye2022don} use factoid questions, hence the participant's alert levels~(reflected in $\beta$ band \citet{55pfurtscheller1999event}) will have big changes if they find the direct answer~(one or two words). 
Unlike their study, the questions in our study are non-factoid and need the participants to judge the document's relevance with more specific content. 
Therefore, another major neurological phenomenon, i.e., valence, may play a major role in our settings, which is also revealed in \textcolor{blue}{existing research~\cite{moshfeghi2013cognition}}.

\subsubsection{Brain Decoding Performance}
\label{Brain Decoding Performance}
The averaged AUC performance for brain responses to landing page contents~($X_l$) is $0.701$ ($SD=0.059$), slightly better than that to snippet content~($X_s$), i.e., $0.690$ ($SD=0.060$). 
Besides, we observe that the overall performance of the personalized model $G_p$~(AUC=$0.691$, $0.681$ for $X_l$, $X_s$) significantly outperforms the generalized model $G_g$~(AUC=$0.670$, $0.603$ for $X_l$, $X_s$). 
This verifies the assumption that $G_p$ outperforms $G_g$ and training a personalized model for each participant is helpful for the classification performance. 
However, when the collected personalized data size is insufficient~(for the first 100 samples), the performance of $G_p$~(AUC=$0.584$) performs worse than the general model $G_g$~(AUC=$0.627$). 
Hence, in the split-by-timepoint splitting protocol, it is reasonable to transform $G$ from the generalized model $G_g$ to the personalized model $G_p$. 
\textcolor{blue}{
Unlike randomized data splitting, our split-by-timepoint splitting protocol closely resembles real-world scenarios to simulate the cold start situation that new users experience when they begin using our system.
The results of our experiment illustrate that using a general model during the cold start phase and training a subject-specific model during the search process can yield practicable results. 
But if some users are unwilling to provide their data for model training, utilizing general models and employing techniques like edge computing to prevent user data from being exposed can still yield satisfactory results.
}

\textcolor{blue}{
As EEG data commonly contain noises, the brain decoding performance is not perfect~\cite{25gwizdka2017temporal,72ye2022don}. 
However, unlike explicit annotations, the collection of EEG signals is real-time and does not interfere with the user's search process, which makes it preferable in practice. 
Furthermore, there is also room for improvement in designing a sophisticated strategy to deal with the data variation problem across different individuals and devising more effective classification models. 
Since this is not the focus of this paper, we leave the study of constructing EEG classification models in interactive search scenarios as future work.
}

\subsubsection{\textcolor{blue}{Sensitivity analysis}}
\label{Sensitivity analysis}
\begin{figure}
    \centering
    \begin{subfigure}[b]{0.3\textwidth}
        \centering
        \includegraphics[width=0.9\linewidth]{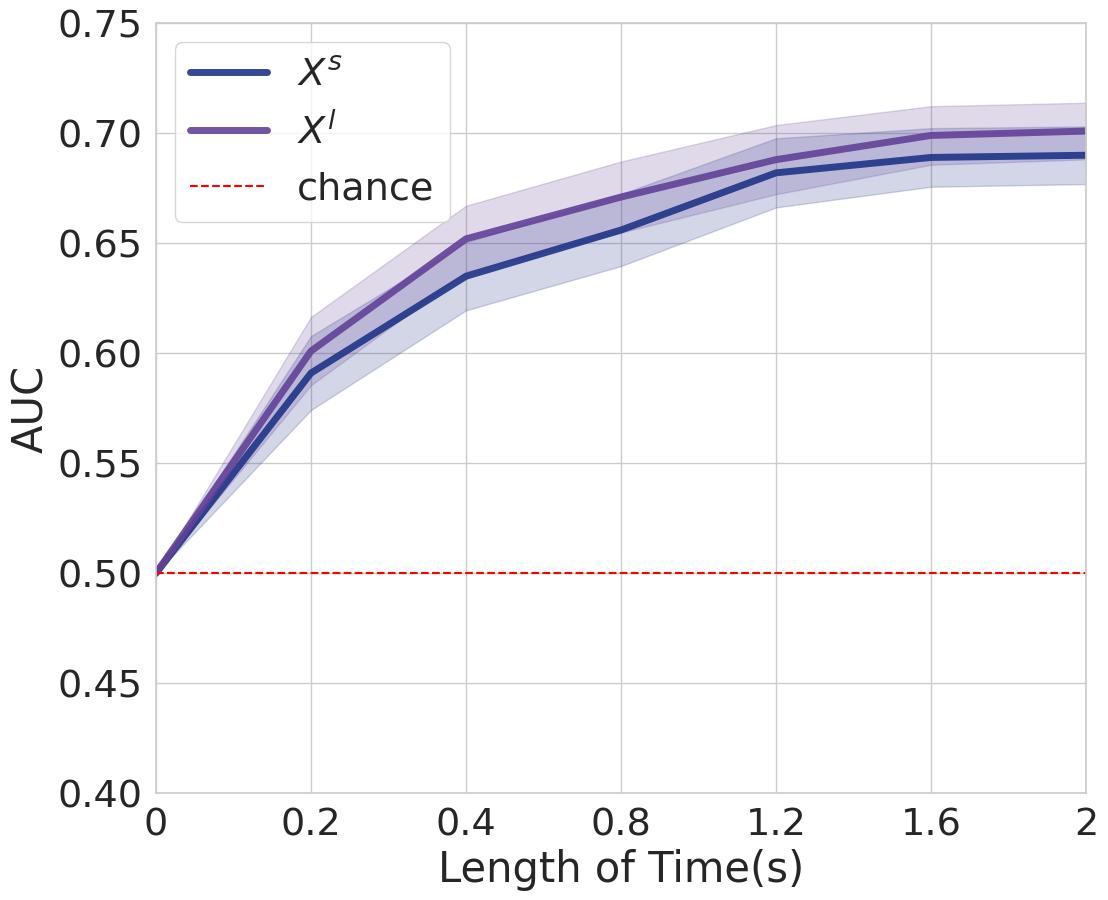}
        \caption{Length of time.}
        \label{fig:time_window}
    \end{subfigure}
    \hfill
    \begin{subfigure}[b]{0.32\textwidth}
        \centering
        \includegraphics[width=0.9\linewidth]{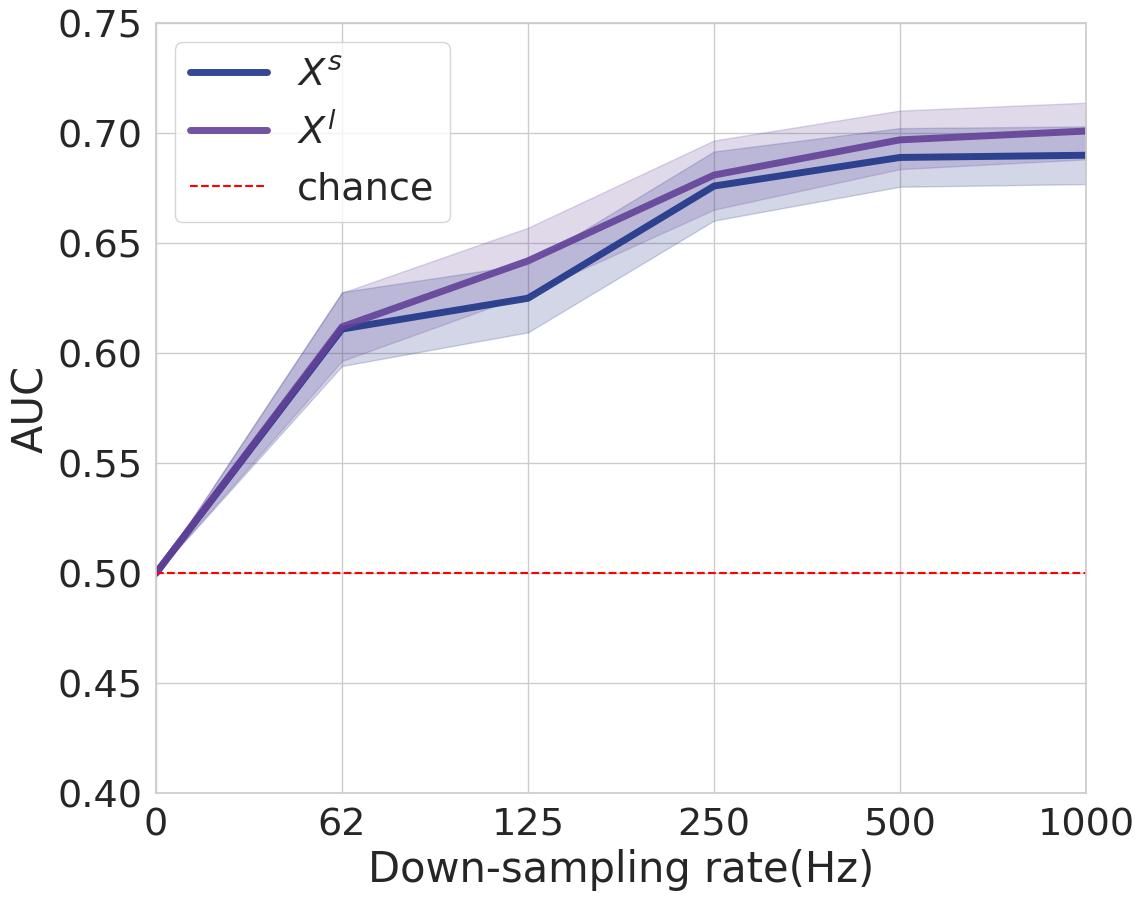}
        \caption{Down-sampling rate.}
        \label{fig:down_sample}
    \end{subfigure}
    \hfill
    \begin{subfigure}[b]{0.3\textwidth}
        \centering
        \includegraphics[width=0.9\linewidth]{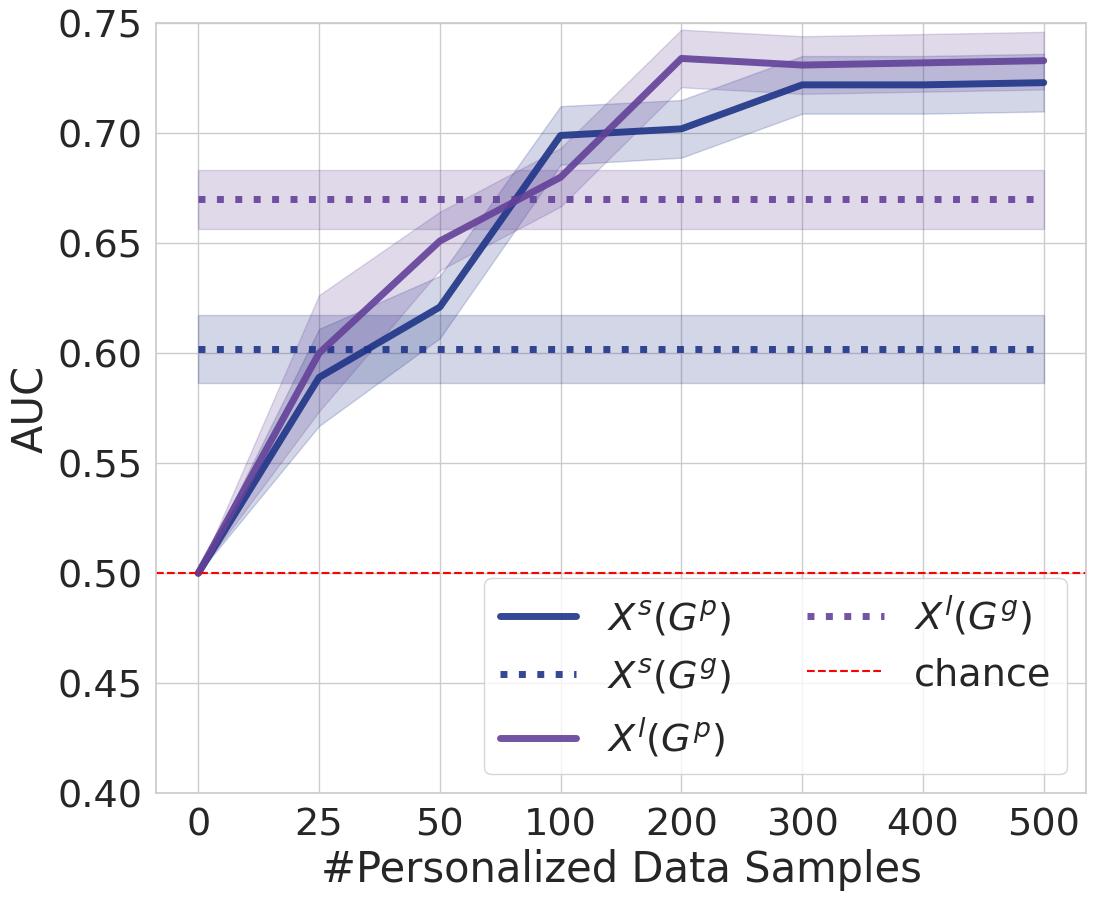}
        \caption{\#Personalized data samples.}
        \label{fig:number_of_data}
    \end{subfigure}
    \caption{\textcolor{blue}{AUC performance for relevance annotations of snippets and landing pages across different lengths of time, down-sampling rate, and number of personalized data samples for training. The shaded regions indicate the standard error.}}
    \label{fig:sensitive}
\end{figure}

\textcolor{blue}{
In Section~\ref{Preprocessing}, we elaborate on the data preprocessing protocols adopted in our study, which follows a common setting with existing neuroscientific studies~\cite{56pinkosova2020cortical,72ye2022don}.
To explore and understand how these settings will affect the relevance prediction performance, we conduct a sensitivity analysis regarding the data-preprocessing protocols, especially the length of time segmentation and down-sampling rate, as shown in 
Figure~\ref{fig:time_window} and Figure~\ref{fig:down_sample}, respectively.
From Figure~\ref{fig:time_window}, we can observe that the relevance estimation performance improves with the time duration increases and stabilizes between 1,600 ms to 2,000 ms.
Existing literature~(\cite{3allegretti2015relevance} and \cite{72ye2022don}) reported that differences in users' brain signals begin to emerge at 800ms for relevant and irrelevant documents. 
However, we find there is still room for improvement with data collected over a longer length of time.
On the other hand, we find that the relevance estimation performance improves if we don't down-sample too much, as shown in Figure~\ref{fig:down_sample}.
To reduce computational complexity, especially in real-time scenarios pertinent to relevance feedback, we adopted a consistent down-sampling rate of 500 Hz. 
This ensures comparable performance to scenarios without down-sampling, yet with fewer computational demands.
}

\textcolor{blue}{
Additionally, we also conducted a sensitivity analysis regarding the number of training personalized data samples.
In this analysis, we train the brain decoding model $G^p$ with different sizes of collected brain data samples $\mathcal{X}$, and evaluate it with the additional 100 data samples.
Furthermore, we undertake a sensitivity analysis pertaining to the number of training personalized data samples.
For each participant, we train the brain decoding model $G^p$, using varying sizes of collected brain data samples $mathcal{X}$, and subsequently assess its performance using an additional 100 data samples.
Figure~\ref{fig:number_of_data} depicts the classification performance in terms of AUC.
As shown in Figure~\ref{fig:number_of_data}, we observe that our model trained with cross-subject data~($G^g$) also exhibits good performance, which means that, we may be able to avoid using user's personal data by training and deploying a global brain signal model for everyone while keeping a reasonable retrieval performance.
Additionally, $G^p$ trained with a small amount of training data still performs well in the experiments~(i.e., 100 data samples), which means that, in practice, we just need to collect a limited amount of user data to train the model, and then deploy the model to user's device to avoid further collection of user's personal data.
These observations illustrate that we could reduce the risk of privacy-related issues in practice by using cross-subject modeling and limited sample modeling when applying BCI devices in IR scenarios.
}

\input{meta/statistical_analysis}

\paragraph{Answer to \textbf{RQ1}.} 
We observe that the relevance of web pages can be inferred by decoding brain signals. This demonstrates the potential of using brain signals as an additional relevance indication to supplement existing relevance signals.

%% file: meta/statistical_analysis.tex
\subsubsection{\textcolor{blue}{Mixed effects analyses}}
\label{A.2}
\textcolor{blue}{
In the data collection procedures, we randomized the task order and the order of the search documents, minimizing the risk of potential confounders. 
Despite these precautions, it remains implausible to entirely eliminate the influence of all confounding variables. 
Here we deliberate upon various confounding factors that could potentially influence the statistical robustness and validity of our analytical observations.
Following existing literature~\cite{72ye2022don}, the confounding factors we considered include: individual difference~($I$), the task rank~($O^t$), the documents rank~($O^d$), and the word number in the documents~($W$), and the image size of the displayed document~($S$). 
A linear mixed model is used for modeling the dependence of brain activities~($X$) measured by EEG spectral powers and the ground truth relevance of documents~($R^{gh}$), which can be specified as:
\begin{equation}
\begin{split}
X = (&\beta_u + i_u)R^{gh} + \beta_w W + \beta_t O^t + \beta_d O^d + \beta_s S + I + \beta_0 + e,
\end{split}
\end{equation}
where $e$ represents the global residual error, $\beta_0$ denotes the global intercept, $\beta_u$, $\beta_w$, $\beta_t$, $\beta_d$, $\beta_s$ denote coefficients corresponding to the effects of the confounding factors we list above. 
$I$ is the individual difference effect and $i_u$ is the coefficient for participant $u$.  
The brain activity $X$, estimated via spectral power as elaborated in Section~\ref{5.1}, and the document relevance, $R^{gh}$, which spans a range from 1 to 4.
}

\begin{table}
\centering
\caption{\textcolor{blue}{The statistical results of the mixed linear model. Coef. and z indicate the coefficient variable and the statistic corresponding to the effect measured by brain activities, respectively.}}
\label{tab:mixed}
\begin{tabular}{lcccc}
\hline
Effects             & Coef. & Std  & z     & p>z   \\
\hline
Word number   & 0.000  & 0.000 & 1.247  & 0.106 \\
Document rank & -0.007 & 0.004 & -2.320 & 0.010 \\
Task rank     & -0.001 & 0.000 & -1.542 & 0.062 \\
Image size    & 0.001  & 0.000 & 0.099  & 0.921 \\
Document relevance  & 0.041 & 0.005 & 8.921 & 0.000 \\
\hline
\end{tabular}
\end{table}

\textcolor{blue}{
Table~\ref{tab:mixed} illustrates the coefficient variables and significance of the fixed effect~(i.e., the document relevance), and the random effects~(i.e., word number, document rank, task rank, and image size).
From Table~\ref{tab:mixed}, we observed that the document relevance exerted a significant effect on the brain response.
This effect stood out prominently when compared to the confounding variables. 
Among the confounding factors, only the document rank demonstrated a significant influence on the outcome, which is reasonably attributable to the position bias effect~\cite{31joachims2017unbiased,5azzopardi2021cognitive}, where users typically perceive documents at the top of the list as more relevant.
Conversely, the remaining confounding factors showed minimal effects that were not statistically significant. 
This suggests that while there are multiple factors influencing the brain response, the relevance of the document plays a predominant role.
}

%% file: 5.2.tex
\subsection{IRF \& RRF Experiment~(RQ2)}
\label{5.2}
\subsubsection{Experimental Setup}
\label{5.2.1}

\textbf{Evaluating Protocols:}
In IRF, we evaluate the re-ranking performance of the upcoming documents $\mathcal{D}_u = \{d_{h+1}, d_{h+2}, \ldots, d_n\}$ in an interactive manner: calculate and average the metric $\Pi(R^{gu}, R^{it})$ as $h$ increases from 1 to $h_{max}$. As we do not require the participants to annotate the relevance for unseen documents $\mathcal{D}_u$, we use the external annotations as the ground truth relevance $R^{gu}$. 
In RRF, we calculate the metric $\Pi(R^{ge}, R^{re})$ of the historical documents $\mathcal{D}_h = \{d_1, d_2, \ldots, d_h\}$ for $h = h_{max}$ and adopt the user's annotation to obtain $R^{ge}$. 
Since the landing page contains more content about the document than the snippet content, the document's ground truth relevance is assigned as the landing page's annotation if the document is clicked. 
Otherwise, if the document is non-clicked, we simply use the annotation of the snippet as a substitution. 
\textcolor{blue}{For ranking-based evaluation metric $\Pi$, we adopt Mean Average Precision~(MAP) and Normalized Discounted Cumulative Gain~(NDCG) at different cutoffs: 1, 3, 5, and 10~\cite{clough2013evaluating}.} 
A two-tailed t-test is then applied to measure the significance of the re-ranking performance achieved by different methods and signals.

\textbf{Baselines:}
\textcolor{blue}{For IRF, the baselines include three re-ranking strategies without any user signals: $\text{BM25}$~\cite{robertson2009probabilistic}, $\text{BERT}(R^p)$~\cite{53nogueira2019passage}~(re-rank according to the BERT re-ranker, equivalent to re-ranking by $R^p$), and $\text{Sogou}$~(using the original ranking in the Sogou search engines).} 
Besides, we report the performance of our proposed BERT-based query expansion method $QE^{F_{\theta^{it}}(R^{bs},R^c,R^p)}$ and its ablations~($QE^{R^p}$, $QE^{F_{\theta^{it}}(R^c,R^p)}$, and $QE^{F_{\theta^{it}}(R^{bs},R^p)}$). 
In addition to the proposed RF method, we also report the performance of a traditional RF method $\text{RM3}$~\cite{38lavrenko2017relevance}. 
The implementation of $\text{RM3}$ inherits parameters from ~\citet{38lavrenko2017relevance} and selects the number of rewriting terms from $\{3, 5, 10\}$.
\textcolor{blue}{For RRF, the baselines also include three re-ranking strategies without user signals: $\text{BM25}$, $\text{BERT}$, and $\text{Sogou}$.} 
In addition to the proposed framework, which combines relevance scores from all signals~($F_{\theta^{re}}(R^{bs},R^c,R^p)$), we also report its ablations $F_{{\theta}^{re}}(R^c,R^p)$~(without brain signals) and $F_{\theta^{re}}(R^{bs},R^p)$~(without click signals).

\textbf{Selection of Combination Parameters:}
Each combination parameter $\theta^{*,\dagger}$~($* \in \{it, re\}$, $\dagger \in \{bs, c, p\}$) is selected from $\{0.0, 0.2, 0.4, 0.6, 0.8, 1.0\}$. 
In our experiments, we first explore the overall document re-ranking performance of our framework with fixed values of $\Theta^{it}$ and $\Theta^{re}$. 
To initialize the fixed parameters, we randomly sample a subset consisting of 200 search tasks and test the document re-ranking performance in terms of NDCG@10. Then the optimal parameters among all the combinations~($\Theta^{it}$ in IRF, $\Theta^{re}$ in RRF) on this subset are selected, which are $\theta^{it,bs:c:p}=3:1:1$ in IRF, and $\theta^{re,bs:c:p}=5:2:0$ in RRF. 
We observe that the chosen parameters also achieve the best performance in the whole dataset in comparison with other parameters, which indicates that the chosen parameters are robust. 
Note that similar to most existing RF frameworks~\cite{57rocchio1971relevance,59ruthven2003survey,74yu2021improving}, our experiment does not involve a large number of training parameters. 
Instead, we only tune the combination parameters to merge different RF signals and use a pre-trained retrieval model for document re-ranking.

\textcolor{blue}{As $\theta^{re,p}$ is 0 in the selection of parameters for RRF, we further search in the 0-0.2 range and evaluate the performance in the whole dataset.
We observe that the performance of $F_{\Theta^{re}}(R^{bs}, R^c, R^p)$ yields no significant difference in the range of 0-0.14~(peaks at 0.06).
However, the performance of $BERT(R^p)$ is significantly higher than chance~(i.e., rank the documents randomly) with $p < 0.001$. 
This indicates that pseudo relevance is effective for RRF, but when compared to the user signals~($R^c$ and $R^{bs}$), the effect is very small. 
As the documents in our experiment are selected from top documents while submitting the query to Sogou's search engine, as explained in Section~\ref{Stimuli Prepartion}.
Therefore, most of them are already semantically related to the query term and their semantic relevance to the query is similar, which is pseudo-relevance measures.
Therefore, knowing what topics the user needs is more important than the little differences in semantic relevance, given that the queries in our tasks are short and with broad topics.
We acknowledge the limitation of this selection of queries and further discuss it in Section~\ref{6.3}. 
}

In addition, we also explore the performance of different combination parameters in Section~\ref{5.3} to show how we should re-weight the importance of various RF signals regarding the search scenarios. 
Motivated by the findings in Section~\ref{5.3}, we further explore an adaptive RF signals combination method in Section~\ref{5.4} which combines the importance of RF signals depending on the search scenario. 
The adaptive signals combination method achieves better performance improvement than using fixed parameters in the IRF task.

\subsubsection{Overall Results}
\input{meta/Table_3}

\textbf{IRF Performance.} Table 3 presents the document re-ranking performance in IRF. From Table 3, we have the following observations:
\textcolor{blue}{(1)~Methods using feedback information gathered from user interactions (i.e., $QE^{F_{\theta^{it}}(R^{bs},R^p)}$, $QE^{F_{\theta^{it}}(R^c, R^p)}$, and $QE^{F_{\theta^{it}}(R^{bs}, R^c, R^p)}$) outperform methods without regard to user signals~(i.e., $\text{BM25}$,$\text{BERT}(R^p)$, $\text{Sogou}$, and $QE^{R^p}$).  
Although existing research underscores the superior performance of BERT over BM25~\cite{53nogueira2019passage}, the divergence in their performance becomes notably minimal when juxtaposed with the impact of integrating additional feedback information, especially brain signals.}
This indicates that personal factors extracted from user interactions are helpful for improving document re-ranking performance.
(2)~With additional relevance score $R^{bs}$ extracted from brain signals, the query expansion~($QE$) method receives a significant performance boost. 
The difference between the performance~(in terms of NDCG@10) of $QE^{F_{\theta^{it}}(R^{bs},R^p)}$~($QE^{F_{\theta^{it}}(R^{bs},R^c,R^p)}$) and $QE^{R^p}$ ($QE^{F_{\theta^{it}}(R^c,R^p)}$) is 5.3\%~(1.5\%), which is significant at $p=7.8\times10^{-16}$~($1.5\times10^{-6}$) with a pairwise t-test. 
This demonstrates that brain signals can provide additional information to existing signals~(pseudo-relevance signals or a combination of pseudo-relevance signals and click signals).
(3)~The proposed BERT-based query expansion method is more effective than the conventional method RM3. 
This may suggest that BERT is better at capturing a document's semantic representation than statistical language models.

\input{meta/Table_4}

\textbf{RRF Performance.} Table 4 presents the document re-ranking performance in RRF. We observe similar findings as we have discussed in IRF. 
\textcolor{blue}{First, models that use user signals (i.e., $F_{\theta^{re}}(R^{bs},R^p)$, $F_{\theta^{re}}(R^c,R^p)$, and $F_{{\theta}^{re}}(R^{bs},R^c,R^p)$) often have better performance than those that do not (i.e., $\text{BM25}$, $\text{BERT}$ and $\text{Sogou}$).}
Second, brain signals can boost RF performance, e.g., $F_{{\theta}^{re}}(R^{bs},R^c,R^p)$ leads to a performance gain of 7.4\% in terms of NDCG@10 over $F_{{\theta}^{re}}(R^c,R^p)$. 
These similar findings indicate that in both IRF and RRF, brain signals can be utilized to improve document re-ranking performance.

Except for the similarities, we also notice differences between IRF and RRF. 
For example, we observe that the performance gain achieved by adding user signals (i.e., brain signals or click signals or their combination) is larger in RRF than in IRF. 
Especially, in RRF, the performance gain between $F_{{\theta}^{re}}(R^{bs},R^c,R^p)$ and $F_{\theta}^{re}(R^c,R^p)$~($F_{{\theta}^{re}}(R^{bs},R^c,R^p)$ and $R^p$) is 7.4\% (37.3\%) in terms of NDCG@10.
However, in IRF, the performance gain between $QE^{F_{\theta^{it}}(R^{bs},R^c,R^p)}$ and $QE^{F_{\theta^{it}}(R^c,R^p)}$ ($QE^{F_{\theta^{it}}(R^{bs},R^c,R^p)}$ and $QE^{R^p}$) is smaller, i.e., 1.5\%~(16.9\%) in terms of NDCG@10. 
One possible reason is the fact that ranking unseen documents is much more difficult than ranking historical documents. 
With accurate RF signals, we can easily create the best ranking of historical documents, but may still struggle in estimating the relevance of new documents because there is no guarantee that the new relevant documents would be similar to the historical relevant documents we observed. 
However, in Section~\ref{5.3}, we compare the potential improvements brought by ideally combining different RF signals and observe a larger potential in IRF. Then in Section~\ref{5.4}, we propose a combination method that helps IRF exhibit comparable performance to RRF.

\paragraph{Answer to \textbf{RQ2}.}
We verify that the RF performance can be significantly improved when adding brain signals into the proposed framework, i.e., a performance gain of 7.4\% and 1.5\% in terms of NDCG@10 in IRF and RRF, respectively.

%% file: meta/Table_3.tex
\begin{table}
\begin{threeparttable}[b]
\caption{The document re-ranking performance in IRF. $*$ indicates a significant difference in performance when compared to $QE^{F_{\Theta^{it}}(R^{bs}, R^c, R^p)}$, significant at a level of $p < 1 \times 10^{-3}$.\label{tab:irf}}
\setlength{\tabcolsep}{3mm}{
\begin{tabular}{@{}l!{\color{lightgray}\vrule}ccccc}
\specialrule{0em}{1pt}{1pt}
\toprule
\textbf{Method}\tnote{1} & \textbf{NDCG@1} & \textbf{NDCG@3} & \textbf{NDCG@5} & \textbf{NDCG@10} & \textbf{MAP} \\ \midrule
BM25 & 0.2031$^*$ & 0.2265$^*$ & 0.2449$^*$ & 0.3031$^*$ & 0.3048$^*$ \\
Sogou & 0.2085$^*$ & 0.2284$^*$ & 0.2481$^*$ & 0.3065$^*$ & 0.3214$^*$ \\
BERT~($R^p$) & 0.2060$^*$ & 0.2326$^*$ & 0.2579$^*$ & 0.3221$^*$ & 0.3033$^*$ \\
$QE^{F_{\Theta^{it}}(R^p)}$ & 0.2306$^*$ & 0.2361$^*$ & 0.2582$^*$ & 0.3205$^*$ & 0.3279$^*$ \\
$QE^{F_{\Theta^{it}}(R^{bs}, R^p)}$ & 0.2477$^*$ & 0.2569$^*$ & 0.2774$^*$ & 0.3374$^*$ & 0.3235$^*$ \\
$QE^{F_{\Theta^{it}}( R^c, R^p)}$ & 0.2842$^*$ & 0.2952$^*$ & 0.3124$^*$ & 0.3690$^*$ & 0.3708$^*$ \\
$RM3^{F_{\Theta^{it}}(R^{bs}, R^c, R^p)}$ & 0.2332$^*$ & 0.2523$^*$ & 0.2717$^*$ & 0.3289$^*$ & 0.3337$^*$ \\
$QE^{F_{\Theta^{it}}(R^{bs}, R^c, R^p)}$ & \textbf{0.2948} & \textbf{0.3024} & \textbf{0.3191} & \textbf{0.3747} & \textbf{0.3744} \\ \bottomrule
\end{tabular}
}
\begin{tablenotes}
\item[1] $R^*$ indicates relevance score based on signals $^*$. $bs$, $c$, and $p$ indicate brain signals, click signals, and pseudo-relevance signals, respectively.
\end{tablenotes}
\end{threeparttable}
\end{table}

%% file: meta/Table_4.tex
\begin{table}
\begin{threeparttable}[b]
\caption{The document re-ranking performance in RRF. $*$ indicates the difference in performance, when compared to $F_{\Theta^{re}}(R^{bs}, R^c, R^p)$, is significant at a level of $p < 1 \times 10^{-3}$.\label{tab:rrf}}
\setlength{\tabcolsep}{3mm}{
\begin{tabular}{@{}l!{\color{lightgray}\vrule}ccccc}
\specialrule{0em}{1pt}{1pt}
\toprule
\textbf{Method}\tnote{1} & \textbf{NDCG@1} & \textbf{NDCG@3} & \textbf{NDCG@5} & \textbf{NDCG@10} & \textbf{MAP} \\ \midrule
BM25 & 0.3113$^*$ & 0.3831$^*$ & 0.4703$^*$ & 0.5587$^*$ & 0.5477$^*$ \\
BERT~($R^p$) & 0.3218$^*$ & 0.3846$^*$ & 0.4626$^*$ & 0.5605$^*$ & 0.5534$^*$ \\
Sogou & 0.3313$^*$ & 0.4046$^*$ & 0.4940$^*$ & 0.5848$^*$ & 0.5633$^*$ \\
$F_{\Theta^{re}}(R^{bs}, R^p)$ & 0.5074$^*$ & 0.5494$^*$ & 0.6254$^*$ & 0.6973$^*$ & 0.6744$^*$ \\
$F_{\Theta^{re}}(R^c, R^p)$ & 0.5426$^*$ & 0.5936$^*$ & 0.6578$^*$ & 0.7161$^*$ & 0.6694$^*$  \\
$F_{\Theta^{re}}(R^{bs}, R^c, R^p)$ & \textbf{0.6350} & \textbf{0.6617} & \textbf{0.7171} & \textbf{0.7693} & \textbf{0.8009} \\ \bottomrule
\end{tabular}
}
\begin{tablenotes}
\item[1] $R^*$ indicates relevance score based on signals $^*$. $bs$, $c$, and $p$ indicate brain signals, click signals, and pseudo-relevance signals, respectively.
\end{tablenotes}
\end{threeparttable}
\end{table}

%% file: 5.3.tex
\begin{figure}
    \hspace*{\fill}%
    \subcaptionbox{IRF performance.\label{fig:udr}}
    {\includegraphics[width=.45\linewidth]{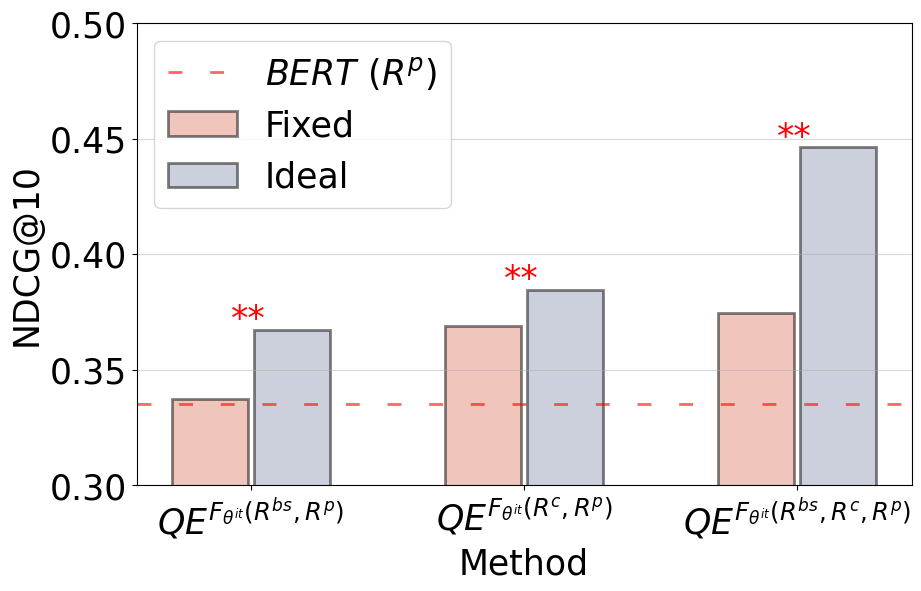}}
    \hfill\hfill\hfill\hfill%
    \subcaptionbox{RRF performance.
    \label{fig:rdr}}
    {\includegraphics[width=.45\linewidth]{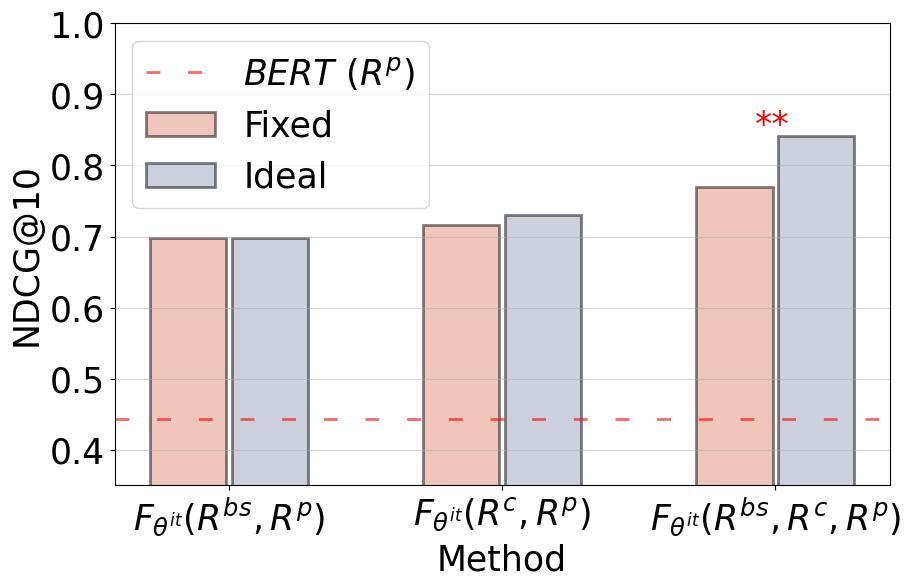}}%
    \hspace*{\fill}%
    \caption{The RF performance with fixed and ideal combination parameter $\Theta$. $**$ indicates significant differences between RF methods using fixed and ideal combination parameters at $p < 0.01$ level using a pair-wise t-test.\label{fig:fixed_ideal}}
\end{figure}

\begin{figure}
    \hspace*{\fill}%
    \subcaptionbox{IRF with \& without $R^{bs}$.\label{fig:udr1}}
    {\includegraphics[width=.45\linewidth]{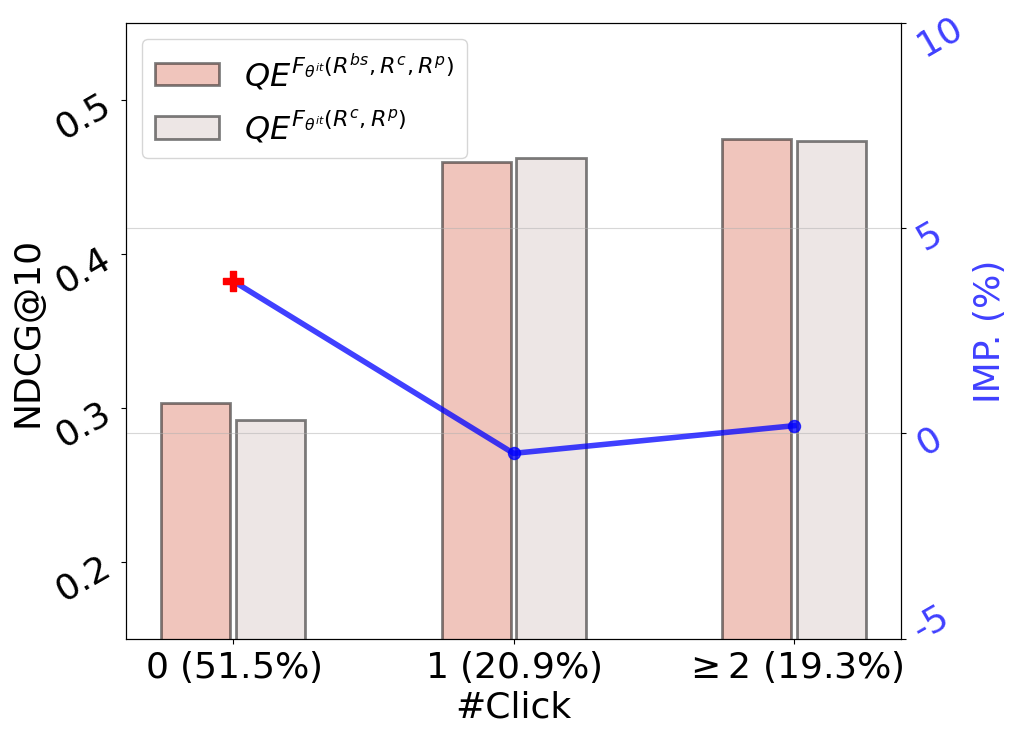}}
    \hfill\hfill\hfill\hfill%
    \subcaptionbox{RRF with \& without $R^{bs}$
    \label{fig:rdr1}}
    {\includegraphics[width=.45\linewidth]{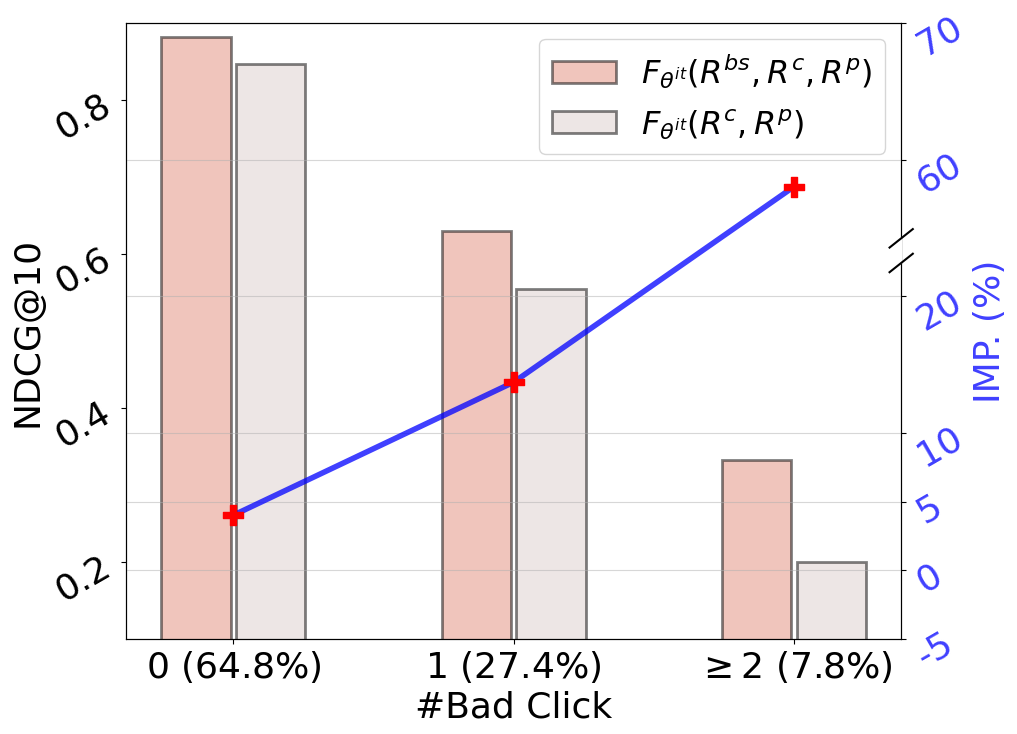}}%
    \hspace*{\fill}%
    \caption{The RF performance with \& without brain signals in different search scenarios. $+$ denotes the improvement~(\textcolor{trueblue}{IMP.}) is significant~($p < 0.05$) using a pairwise t-test.\label{fig:brain_ablation}}
\end{figure}

\subsection{In-depth Analyses~(RQ3)}
\label{5.3}
This section explores the RF framework based on \textbf{combination parameter analyses} and \textbf{search scenario analyses}. 
The combination parameter analyses utilize an ideal experiment to examine the potential of improving RF performance by selecting ideal combination parameters in different scenarios. 
It is observed that potential advancement lies in designing better combination weights for RF signals, especially in IRF. 
On the other hand, the search scenario analyses present the gain of brain signals on RF performance in several specific search scenarios. 
The analyses further show that it is necessary to increase the combination weight of brain signals in those search scenarios. 
The analyses in these two aspects reveal that RF performance improvement lies in designing better combination protocols, which is elaborated in Section~\ref{5.4}.

\subsubsection{Combination Parameter Analyses}
\label{5.3.1}
In the above experiment, we used a fixed selection of combination parameters $\Theta^{it}$ and $\Theta^{re}$ which has averagely the best performance~(detailed in Section~\ref{5.2.1}). 
However, there is still potential for improvement in using different parameters to combine RF signals for each data sample. 
Hence, we conduct an ideal experiment, which adaptively selects the ideal combination parameters $\Theta^{it}$ and $\Theta^{re}$  to achieve the best document re-ranking performance. 
\textcolor{blue}{For every data sample, corresponding to each computation in Algorithm~\ref{algorithm:pipeline}~(line 10 utilizing $\Theta^{it}$ and line 16 employing $\Theta^{re}$), we search values for $\theta^{,\dagger}$, where $ \in {it, re}$ and $\dagger \in {bs, c, p}$, from the set ${0.0, 0.2, 0.4, 0.6, 0.8, 1.0}$. 
Subsequently, we determine the optimal combination of parameters $\Theta^{it}$ and $\Theta^{re}$ based on the NDCG@10 for each distinct data sample.
Note that the performance of the ideal experiment is not achievable in reality. 
This ideal experiment aims to compare the difference between the practical experiment results and the ideal results, which indicates the potential of the RF framework and brain signals.
}

Figure~\ref{fig:fixed_ideal} presents the RF performance with fixed and ideal combination parameter $\Theta$. 
From Figure~\ref{fig:fixed_ideal}, we observe that RF with the ideal combination parameters outperforms RF with the fixed combination parameters in all methods in both IRF and RRF tasks. 
This is obvious since the selection of the ideal combination parameter is directly based on the RF performance. 
Besides, we observe that the performance difference of RF with the fixed/ideal combination parameters is larger in IRF than in RRF. Especially, the performance improvement in IRF achieves 19.1\%~(0.4463 in comparison with 0.3747 in terms of NDCG@10) with the method of $QE^{F_{\theta^{it}}(R^{bs},R^c,R^p)}$, which is significant at $p < 1 \times 10^{-2}$. 
\textcolor{blue}{This indicates that IRF exhibits greater potential than RRF when effectively combining brain signals, click signals, and pseudo-relevance signals.}

A potential explanation for this difference between IRF and RRF is their different task formulation and measurements. 
In RRF, the combination score is directly applied to re-rank the feedback documents~(denoted as $\mathcal{D}_h$ in the task formulation). 
Hence, the combination score only needs to reflect the differences between each other and can be used for ranking purposes. 
However, in IRF, the combination score is applied to a query expansion module. 
Therefore, it is not only necessary to maintain the ranking of documents’ relevance, but also to have reasonable values determine their weight in query rewriting. 
For example, we assume that $d_1$ and $d_2$ are two feedback documents and $d_1$ is more relevant to the user’s intent. 
In RRF, it is no problem to assign any relevance score to $d_1$ and $d_2$ which only needs to maintain that the score of $d_1$ is higher than that of $d_2$. 
However, in IRF, relevance scores of $d_1$ and $d_2$ directly reflect the importance of $d_1$ and $d_2$ in query expansion. 
Hence setting proper relevance scores for them is much more important to avoid possible bad cases, e.g., $d_2$ may even be harmful to query expansion.

\begin{figure}
    \hspace*{\fill}%
    \subcaptionbox{IRF with different $\Theta^{it}$.\label{fig:parameter_udr}}
    {\includegraphics[width=.45\linewidth]{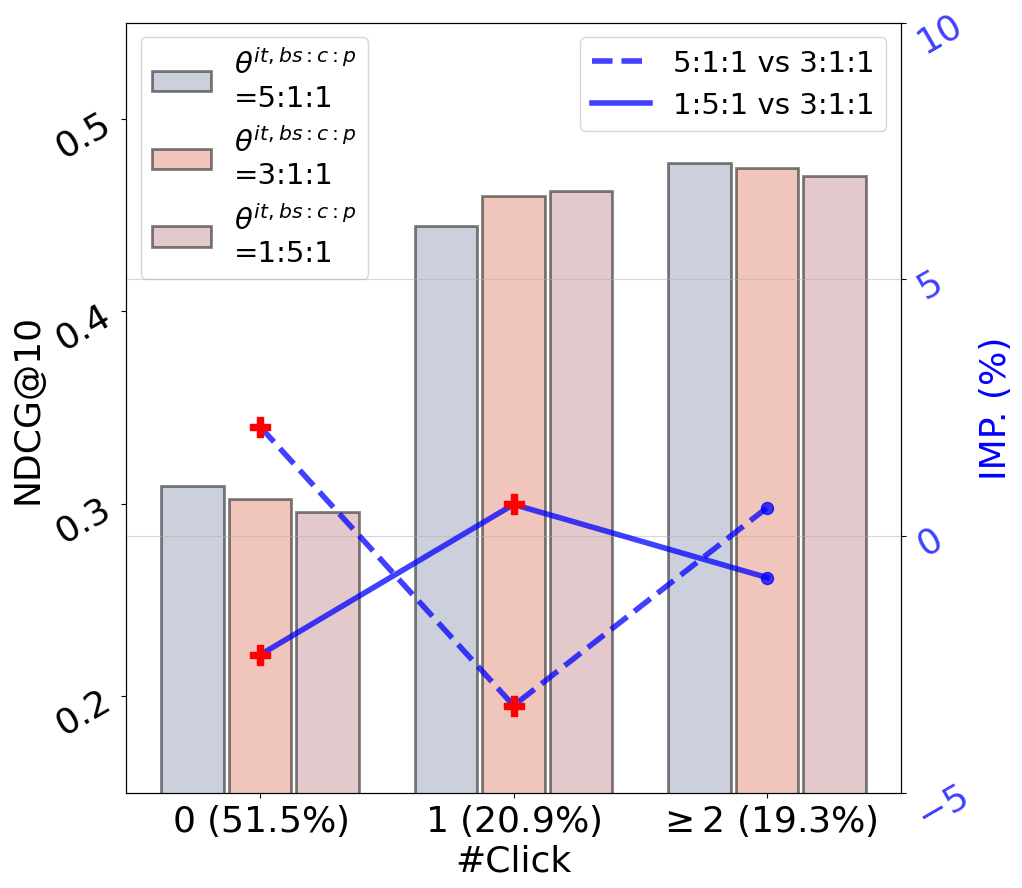}}
    \hfill\hfill\hfill\hfill%
    \subcaptionbox{RRF with different $\Theta^{re}$.
    \label{fig:parameter_rdr}}
    {\includegraphics[width=.45\linewidth]{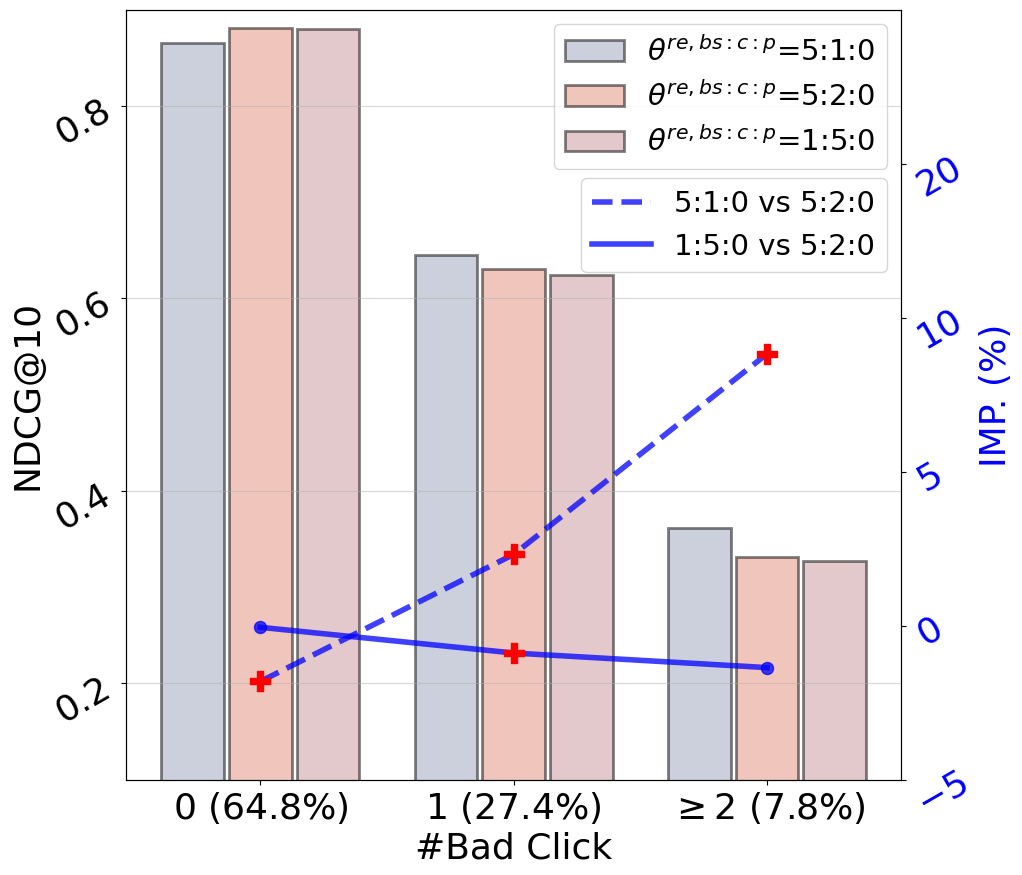}}%
    \hspace*{\fill}%
    \caption{The RF performance with different combination parameter $\Theta$ in various search scenarios. $+$ denotes the improvement~(\textcolor{trueblue}{IMP.}) is significant~(p < 0.05).\label{fig:parameter}}
\end{figure}

\subsubsection{Search Scenario Analyses}
The RF performance in search scenarios where click signals are missing or biased is explored in this section. 
It is observed that brain signals are particularly beneficial when clicks are missing~(analyzed in IRF as kicking off RF before any click happens is significant in an interactive process) by estimating non-click relevance. 
Besides, brain signals help cases where bad clicks happened~(analyzed in RRF so that we can re-rank bad clicks for potential search processes in the future). 

\textbf{Non-click Scenarios.} 
There are good reasons not to treat all non-clicks equally, e.g., some non-clicks may contain useful information~\cite{72ye2022don} and even attract revisiting behaviors~\cite{67xu2012incorporating}. In our user study, we also observe that the non-click documents are annotated with different relevance scores~(1~(77.6\%), 2~(7.4\%), 3~(4.6\%), and 4~(10.2\%)). 
Based on the above observation, we aim to explore whether brain signals can estimate non-click relevance and kick off IRF before any click happens. 
In the brain decoding experiment, we observe that the AUC performance for non-click data samples achieves 0.675, which presents a slight decrease compared to that measured in all data samples but is still significantly better than random. 
This motivates us to further explore whether brain signals can improve RF performance when click signals are missing.

Figure~\ref{fig:udr1} presents the IRF performance in search scenarios with different numbers of clicks. From Figure~\ref{fig:udr1}, we observe that the performance gain between $QE^{F_{\theta^{it}}(R^{bs},R^c,R^p)}$ and its ablation $QE^{F_{\theta^{it}}(R^c,R^p)}$ is largest in a non-click search scenario (an improvement of 3.7\% in terms of NDCG@10). 
In contrast, their performance differences are not significant in search scenarios with not less than one click. 
As IRF is an interactive process, the number of clicks increases as the number of examined documents increases~(Pearson’s $r = 0.45$ ($p < 1 \times 10^{-3}$)). 
Hence, we also observe that the performance difference between $QE^{F_{\theta^{it}}(R^{bs},R^c,R^p)}$ and $QE^{F_{\theta^{it}}(R^c,R^p)}$ is larger at the beginning of the search process (an improvement of 3.3\% in terms of NDCG@10 for session length $h \leq 4$) than the subsequent search process (an improvement of 0.6\% in terms of NDCG@10 for session length $h > 4$). 
As the non-click search scenario accounts for a large proportion~(51.5\%) and it is usually more important to quickly improve retrieval performance at the beginning of the search process, we argue that brain signals have great potential in IRF.

Furthermore, we analyze how the IRF performances vary with the selections of combination parameter $\Theta^{it}$. 
We select $\theta^{it,bs:c:p}$ from $\{3\text{:}1\text{:}1, 5\text{:}1\text{:}1, 1\text{:}5\text{:}1\}$, where 3:1:1 is the averagely optimal parameter among all data samples while 5:1:1 and 1:1:5 are combinations that emphasize the importance of brain signals and click signals, respectively. 
As shown in Figure~\ref{fig:parameter_udr}, in a non-click search scenario, combining brain signals with a higher weight~($\Theta^{it,bs:c:p} = 5\text{:}1\text{:}1$) outperforms other combination parameters $\Theta^{it,bs:c:p} = 3\text{:}1\text{:}1$~($1\text{:}5\text{:}1$) significantly at $p=2.6 \times 10^{-10}$~($6.3 \times 10^{-17}$) using a pair-wise t-test. 
Besides, if we simply set the parameter $\Theta^{it}$ as 5:1:1 in a non-click scenario, and 3:1:1 in others, the IRF performance in terms of NDCG@10 can achieve 0.3781, significantly performing better than the performance achieved by using fixed combination parameters~($p = 2.7 \times 10^{-10}$). 
This reveals the potential benefit if we adaptively integrate brain signals regarding the search context.

\textbf{Bad Click scenarios.}
Bad click indicates a document clicked by the user is irrelevant and may lead to a poor search experience. This usually happens if the document’s snippet is attractive, but its landing page content is unsatisfactory~\cite{45lu2018between}. 
In our experiment, we define a ``bad click'' if the relevance annotation of the landing page is 1~(``totally irrelevant'') or 2~(``irrelevant'') for a clicked document, and we observe that a proportion of 21.8\% clicks are grouped into ``bad click''. 
Since brain signals are effective in inferring the relevance of the landing pages~(with a binary classification AUC=0.703), it is interesting to further explore whether brain signals can boost RRF performance in cases where a bad click happens.

As presented in Figure~\ref{fig:rdr1}, we observe that the performance difference between $F_{\theta^{re}}(R^{bs},R^{c},R^p)$ and $F_{\theta^{re}}(R^{c},R^p)$ is larger as the number of bad clicks increase. This indicates brain signals can bring more benefits in scenarios where bad clicks often happen. 
Furthermore, we explore combination parameters $\Theta^{re}$ that averagely performs best ($\theta^{re,bs:c:p}=5\text{:}2\text{:}0$) and emphasize the importance of brain signals ($\theta^{re,bs:c:p}=5\text{:}1\text{:}0$) and click signals ($\theta^{re,bs:c:p}=1\text{:}5\text{:}0$), respectively. 
From Figure~\ref{fig:parameter_rdr}, we observe that combination with $\theta^{re,bs:c:p}=5\text{:}1\text{:}0$ performs much better than using the averagely optimal parameter $\theta^{re,bs:c:p}=5\text{:}2\text{:}0$. 
This emphasizes the need to prioritize brain signals in cases where there may be negative or improper clicks. 
If we simply adopt parameter $\theta^{re,bs:c:p}=5\text{:}1\text{:}0$ in search scenarios where at least one bad click happens and $5\text{:}2\text{:}0$ in others, the RRF performance in terms of NDCG@10 can achieve 0.7756~(significantly better than using fixed parameters at $p=1.0\times 10^{-3}$ using a pair-wise t-test). 
Although the number of bad clicks is actually not available in practice, this observation reveals a possibility to better combine brain signals into RF. 
The studies of detecting search scenarios that potentially lead to bad click and designing adaptable combination strategies are left as future work.



\subsubsection{\textcolor{blue}{Case study}}
\label{Case studies}
\input{meta/case_study2}

\textcolor{blue}{
Table~\ref{tab:case_study} presents an example of IRF and RRF results for participant ID 1 and query ``The Prophet'' with task description ``explore the concepts of prophet in the general domain''.
In this scenario, the participant goes through historical documents ranging from d1 to d6. 
For the IRF task, the objective is to re-rank the unseen documents~($D^u$), d7 through d12, whereas the RRF task focuses on re-ranking the historical documents($D^h$) from d1 to d6.
Table~\ref{tab:split_table1} presents the document titles.
Under the task description, document $d4$, $d6$, and $d11$ are highly relevant~(with relevance annotation 4). 
On the other hand, documents $d2$ and $d8$ are partially relevant since they are related to the prophet in Islam and Christian, respectively. 
Both are subsets of the broader concept of the prophet.
Documents $d12$ is also partially relevant since the dictionary explanation of ``Prophet'' may provide some useful information,  though it is not concrete enough.
}

\textcolor{blue}{
Table~\ref{tab:split_table2} presents the estimated base relevance scores for historical documents $D^h$.
From Table~\ref{tab:split_table2}, we can observe that $R^p$ does not align with the ground truth relevance very well, as it is only based on the query terms encompassing broad topics.
On the other hand, $R^c$ and $R^{bs}$ are aligned with the ground truth relevance.
Additionally, $R^{bs}$ can provide additional information especially when two documents are both non-click or clicked.
As shown in Table~\ref{tab:split_table3}, in the RRF task, d1,d2,d3, and d5 are non-clicks.
But RF with $R^{bs}$ accurately ranks d2 at the top of them because the user's brain response provides additional information of non-clicks. 
On the other hand, the user's brain response provides information that the user is more satisfied with d6 regarding the clicked documents d4 and d6.
Hence, in the IRF task, RF with $R^{bs}$ ranks d11 ahead of d12 because d11 is semantically more close to the most satisfied historical document~(i.e., d6) than d12.
}

\paragraph{Answer to \textbf{RQ3}.} 
We demonstrate that brain signals are particularly helpful in search scenarios involving non-click relevance estimation~(in IRF) and bad click identification~(in RRF). Therefore, it is imperative to prioritize brain signals in search scenarios where click signals may be biased.

%% file: meta/case_study2.tex
\newlength{\dlength}
\setlength{\dlength}{3.3cm}

\begin{table} 
\caption{\textcolor{blue}{Example of IRF and RRF results for participant ID 1 and query ``The Prophet'' with task description ``explore the concepts of prophet in the general domain''. Documents with relevance annotations of 4 and 2-3 are highlighted in \textcolor{myPurple}{purple} and \textcolor{myLightPurple}{light purple}, respectively. Documents with annotations of 1 are presented in \textcolor{black}{black}.}}
  \label{tab:case_study}
  \centering 
  \begin{minipage}{0.5\linewidth} 
    \centering
    \subcaption{\textcolor{blue}{The titles~(translated into English) for historical documents~($D^h$) and unseen documents~($D^u$).}}
    \label{tab:split_table1}
    \input{meta/split_table1}
  \end{minipage}
  \hfill 
  \begin{minipage}{0.45\linewidth}
    \begin{minipage}{\linewidth}
        \centering
        \subcaption{\textcolor{blue}{The estimated base relevance scores for historical document $D^h$, $R^c$, $R^p$, and $R^bs$ represent click-based, pseudo-relevance based, and brain-based scores, respectively.}}
        \label{tab:split_table2}
        \input{meta/split_table2}
    \end{minipage}
    \vspace{1cm}
    \vfill
    \begin{minipage}{\linewidth}
        \centering
        \subcaption{\textcolor{blue}{The re-ranked documents list for IRF \& RRF tasks across RF models with and without brain signals~(w $R^{bs}$ and w/o $R^{bs}$).}}
        \label{tab:split_table3}
        \input{meta/split_table3} 
    \end{minipage}
  \end{minipage}
\end{table}

%% file: meta/split_table1.tex
\begin{tabular}{|p{\dlength}p{\dlength}|}
\hline
\multicolumn{2}{|c|}{\textbf{Query:  The Prophet}} \\ \hline
\multicolumn{1}{|p{\dlength}|}{\makecell[c]{Historical documents \\ ($D^h$)}}& \makecell[c]{Unseen documents \\ ($D^u$)} \\ \hline
\multicolumn{1}{|p{\dlength}|}{d1: The Prophet - French, rough, crime}             & d7: Analysis and translation of the Prophet's poem  \\ \hline
\multicolumn{1}{|p{\dlength}|}{\textcolor{myLightPurple}{d2: Miracles of the Mohammed Prophet}}  & \textcolor{myLightPurple}{d8: What does the prophet mean in the Bible? - Sogou ask}               \\ \hline
\multicolumn{1}{|p{\dlength}|}{d3: The Prophet poem by Gibran} & d9:  The Prophet movie: free  online resource \\ \hline
\multicolumn{1}{|p{\dlength}|}{\textcolor{myPurple}{d4: The Prophet - Sogou encyclopedia}} & d10: \textit{The Prophet} : Gibran \\ \hline
\multicolumn{1}{|p{\dlength}|}{d5: Fifth personality -   introduction of the Prophet} & \textcolor{myPurple}{d11: The Prophet in different cultures} \\ \hline
\multicolumn{1}{|p{\dlength}|}{\textcolor{myPurple}{d6: General concepts of ``Prophet''}} &  \textcolor{myLightPurple}{d12:  The Prophet - Chinese dictionary}      \\ \hline
\end{tabular}

%% file: meta/split_table2.tex
\begin{tabular}{|llll|}
\hline
$D^h$ & \multicolumn{1}{|l|}{$R^c$} & \multicolumn{1}{l|}{$R^p$} & $R^{bs}$\\ \hline
d1 & \multicolumn{1}{|l|}{\makecell[c]{0}} & \multicolumn{1}{l|}{\makecell[c]{0.6}}& \makecell[c]{ 0.3}    \\ \hline
\textcolor{myLightPurple}{d2} & \multicolumn{1}{|l|}{\makecell[c]{0}}  & \multicolumn{1}{l|}{\makecell[c]{0.3}}  &               \makecell[c]{0.6}            \\ \hline
d3 &  \multicolumn{1}{|l|}{\makecell[c]{0}}& \multicolumn{1}{l|}{\makecell[c]{0.4}}& \makecell[c]{0.3} \\ \hline
\textcolor{myPurple}{d4} &  \multicolumn{1}{|l|}{\makecell[c]{1}} & \multicolumn{1}{l|}{\makecell[c]{0.4}}   &  \makecell[c]{0.7}      \\ \hline
 d5 & \multicolumn{1}{|l|}{\makecell[c]{0}}              & \multicolumn{1}{l|}{\makecell[c]{0.3}}               &       \makecell[c]{0.2}                      \\ \hline
\textcolor{myPurple}{d6} & \multicolumn{1}{|l|}{\makecell[c]{1}}              & \multicolumn{1}{l|}{\makecell[c]{0.5}}               &   \makecell[c]{0.6}                          \\ \hline
\end{tabular}

%% file: meta/split_table3.tex

\begin{tabular}{|l|l|l|}
\hline
\multirow{2}{*}{\textbf{IRF}} & w $R^{bs}$ & \textcolor{myPurple}{d11},\textcolor{myLightPurple}{d12},\textcolor{myLightPurple}{d8},d7,d9,d10 \\ \cline{2-3} 
                     &  w/o $R^{bs}$   & \textcolor{myLightPurple}{d12},\textcolor{myPurple}{d11},\textcolor{myLightPurple}{d8},d7,d9,d10  \\ \hline
\multirow{2}{*}{\textbf{RRF}} &  w $R^{bs}$ & \textcolor{myPurple}{d4},\textcolor{myPurple}{d6},\textcolor{myLightPurple}{d2},d1,d3,d5 \\ \cline{2-3} 
                     &  w/o $R^{bs}$ & \textcolor{myPurple}{d6},\textcolor{myPurple}{d4},d1,d3,\textcolor{myLightPurple}{d2},d5 \\ \hline
\end{tabular}
	

%% file: 5.4.tex
\subsection{Adaptive Combination of RF Signals~(RQ4)}
\label{5.4}
In Section~\ref{5.3}, we have demonstrated that the importance of different RF signals varies with the search scenarios. 
This indicates the potential for an adaptive RF signal combination strategy. Particularly, as shown in Figure 6, IRF has a larger improvement space than RRF since the performance difference between the fixed signal weighting and the ideal signal weighting is huge. 
Motivated by these observations, we propose and experiment with an adaptive RF signals combination method by synthesizing user signals to learn adaptive combination weights in different search scenarios.

\begin{figure}
    \hspace*{\fill}%
    \subcaptionbox{Query ``prophet''.\label{fig:c1}}
    {\includegraphics[width=.31\linewidth]{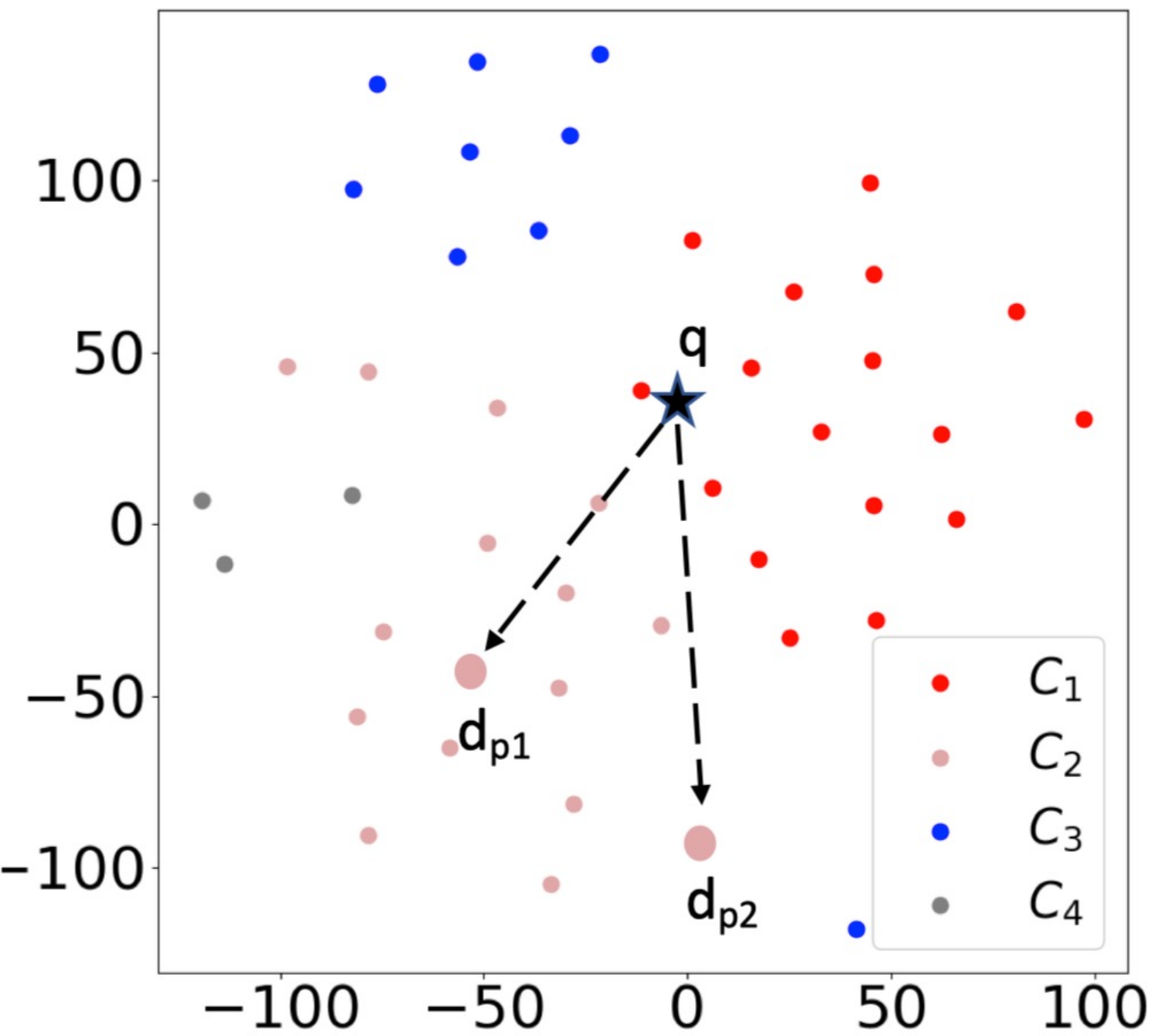}}
    \hfill\hfill\hfill\hfill%
    \subcaptionbox{Query ``tea''.\label{fig:c2}}
    {\includegraphics[width=.31\linewidth]{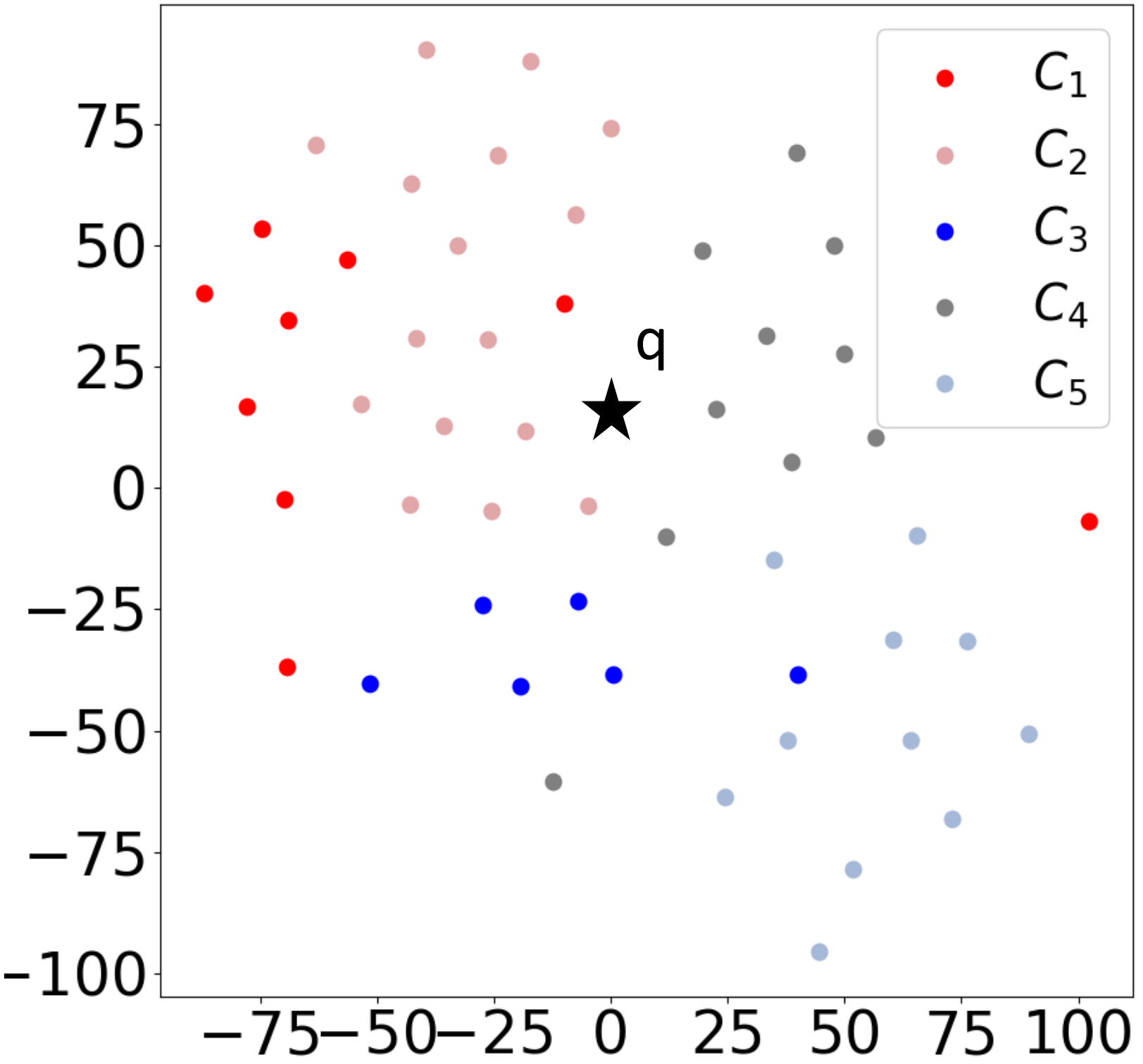}}
    \hfill\hfill\hfill\hfill%
    \subcaptionbox{Query ``wedding ring''.
    \label{fig:c3}}
    {\includegraphics[width=.31\linewidth]{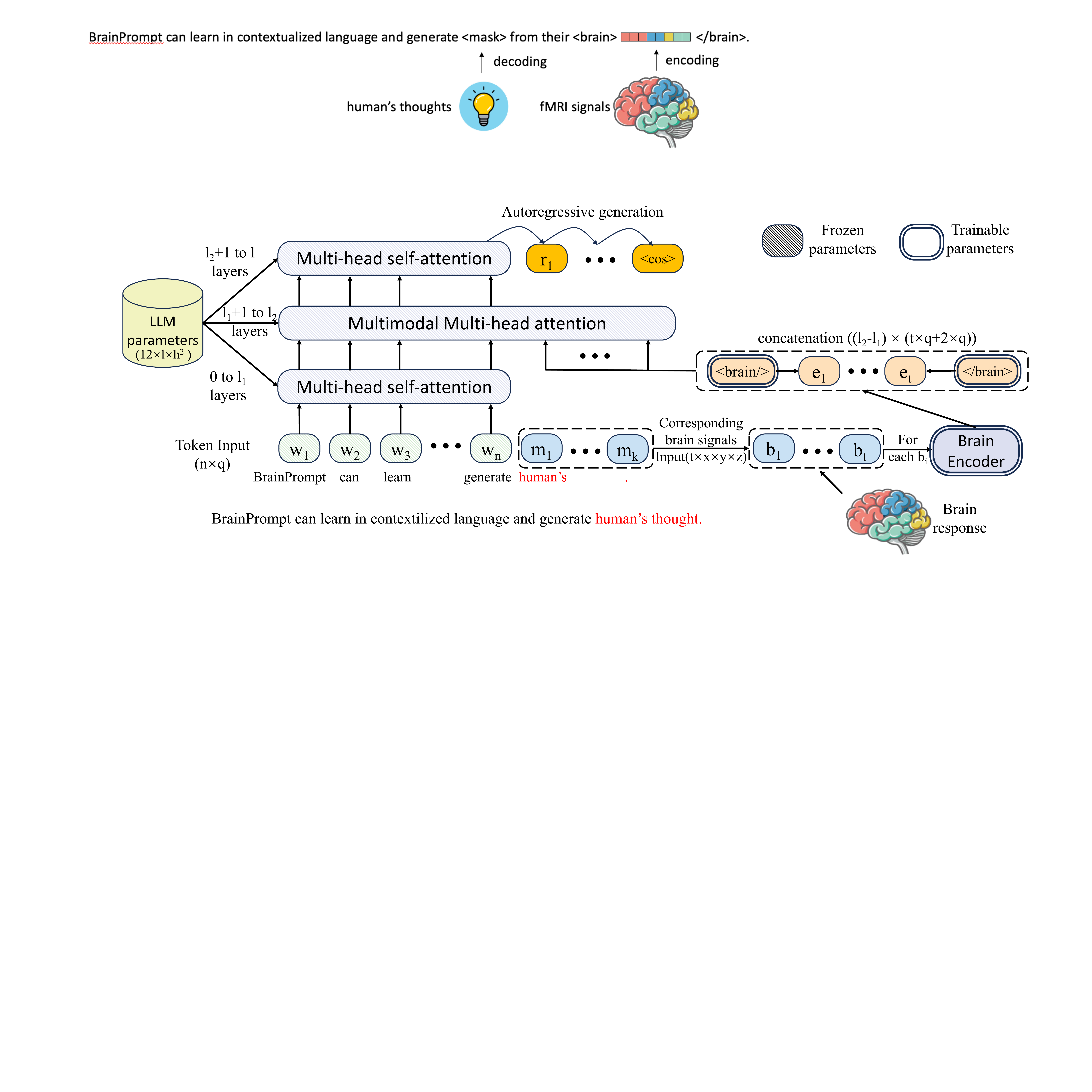}}%
    \hspace*{\fill}%
    \caption{T-SNE plot of BERT embeddings of queries and their corresponding documents with different intents~(painted in different colors). The IRF process utilizes a query expansion module to bring the query representation closer to the user’s current search intent.\label{fig:cluster}}
\end{figure}

\subsubsection{Modeling Motivation}
We illustrate and summarize the modeling motivation of the adaptive RF signals combination method. 
To facilitate ease of understanding, this section inherits the notations presented in Table~\ref{tab:notations}.

As discussed in Section~\ref{5.3}, setting proper combination parameters is more significant in IRF than in RRF. 
Hence, we focus on applying the adaptive RF signals combination method in IRF in this section. 
Before introducing the motivation of the adaptive RF signals combination method, we present an illustration of the IRF task in Figure 9. 
The IRF process is designed to bring the query representation closer to the user’s current search intent, as queries submitted to the search engine are usually broad and may be related to different subtopics. 
For instance, as depicted in Figure 9a, the documents corresponding to the query ``prophet'' are categorized into four clusters $\mathcal{C}_1$, $\mathcal{C}_2$, $\mathcal{C}_3$, and $\mathcal{C}_4$, each representing a different subtopic. 
Suppose the user’s search intent is linked to the documents in cluster $\mathcal{C}_2$ and shows a high RF score on document $d_{p1}$. 
In that case, the IRF will bring the original query representation closer to $d_{p1}$. 
Thus, other documents~(e.g., $d_{p2}$) in cluster $\mathcal{C}_2$ would acquire higher relevance scores than documents in other clusters.

We denote a possible search scenario as $\mathcal{S}_c = \{q, \mathcal{D}_h, \mathcal{D}_u, n_h\}$ in which the user has examined $h$ documents under query $q$, and the number of user’s clicks is $n_h$~($n_h \in \{1, 2, ..., h\}$). 
Then we explain why the combination parameter $\Theta^{it, \mathcal{S}_c} = \{\theta^{it, bs, \mathcal{S}_c}, \theta^{it, c, \mathcal{S}_c}, \theta^{it, p, \mathcal{S}_c}\}$ should be specially selected for different search scenarios $\mathcal{S}_c$. 
First, to effectively utilize the documents in $\mathcal{D}_h$ as feedback signals, the documents’ representativeness within their respective clusters should be considered. For example, as shown in Figure 9a, both document $d_{p1}$ and document $d_{p2}$ belong to the same document cluster $\mathcal{C}_2$. 
However, $d_{p1}$ is more representative than $d_{p2}$ as it is closer to the average representation of documents in $\mathcal{C}_2$. 
Hence, the feedback information of $d_{p1}$ is more valuable than $d_{p2}$ as it is more representative among documents in $\mathcal{C}_2$. 
To boost RF performance, when $d_{p1}$~($d_{p2}$) belongs to the feedback documents $\mathcal{D}_h$, $\Theta^{it, \mathcal{S}_c}$ should be assigned with a higher~(lower) weight. 
Second, the click probability of a document is influenced not only by its relevance but also by other independent factors, e.g., some irrelevant documents may attract users’ clicks due to the ``clickbait'' issue. 
Hence, whether the click signals are reliable and how should we balance the combination weights of click signals and brain signals~(i.e., balance $\theta^{it, bs, \mathcal{S}_c}$ and $\theta^{it, c, \mathcal{S}_c}$) should depend on the search scenario $\mathcal{S}_c$. 
Finally, the distance between documents in $\mathcal{D}_h$ and the original query $q$ varies with different documents. For example, as shown in Figure 9c, $d_y$ is more similar to the original $q$ than $d_g$. Hence, the weight of $\theta^{it, p, \mathcal{S}_c}$ should be set differently depending on the representation difference between the documents in feedback documents $\mathcal{D}_h$ and the original query $q$.

\subsubsection{Overall Pipeline}
As shown in Algorithm~\ref{algorithm}, the adaptive RF signals combination method generates adaptive combination parameters $\Theta^{it, \mathcal{S}_c}$ for a possible search scenario $\mathcal{S}_c = \{q, \mathcal{D}_h, \mathcal{D}_u, n_h\}$. 
The method includes the following process: 
First, we cluster the documents $\mathcal{D}$ corresponding to the query $q$ into $q_m$ clusters $\{\mathcal{C}_1, \mathcal{C}_2, ..., \mathcal{C}_{q_m}\}$. 
Second, for each document cluster $N$, we assume that documents within this cluster are related to the user’s search intent and synthesize the user’s possible click-based and brain-based relevance scores $\{R^{c, \mathcal{S}_c}, R^{bs, \mathcal{S}_c}\}$ under this assumption. 
Finally, with the synthesized user signals $\{R^{c, \mathcal{S}_c}, R^{bs, \mathcal{S}_c}\}$, we search for the combination parameters $\Theta_{it}$ with the averagely best RF performance, which is denoted as $\Theta^{it, \mathcal{S}_c}$. With the above process, we generate the adaptive RF combination parameters $\Theta^{it, \mathcal{S}_c}$ that can be applied to combine the actual user signals during the search process.

\input{meta/algorithm_1.tex}

\subsubsection{Preparation}
We cluster the documents $\mathcal{D}$ corresponding to the query $q$ into $q_m$ clusters $\{\mathcal{C}_1, \mathcal{C}_2, ..., \mathcal{C}_{q_m}\}$ following Liu et al. [43]’s method. Liu et al. [43] assume that a query may contain several subtopics and adopt a multi-step procedure to generate the subtopics and classify each document into one of the subtopics with manual effort. Examples of the document clustering results are presented in Figure 9, in which each document is visualized with its BERT embeddings. As the document clustering method is not the concentration of this paper, we leave the exploration of other automatic subtopic mining baselines with competitive performance or unsupervised settings as future work.

\subsubsection{Click-based and Brain-based Relevance Score Synthesis}
For a search scenario $\mathcal{S}_c = \{q, \mathcal{D}_h, \mathcal{D}_u, n_h\}$, we synthesize the user’s possible click-based and brain-based relevance scores $R_{c, \mathcal{S}_c}$ and $R^{bs, \mathcal{S}_c}$ for each document in $\mathcal{D}_h$. 
We assume that query $q$ is related to $q_m$ search intents $\{Si_1, ..., Si_{q_m}\}$ in which $Si_{j}$ is related to one of the cluster $\mathcal{C}_j$ ($j \in \{1, \ldots, q_m\}$). 
Then we iterate over every cluster $\mathcal{C}_j$ and synthesize $N$~(set as 20) user behaviors on each iteration. 
The synthesis will generate click-based relevance scores $R^{c,Sc}$ and brain-based relevance scores $R^{bs,Sc}$ for each scenario $Sc$. 
Note that we simply assume that all search intents $Si_{j}$ corresponding to a given query have uniform possibilities. 
\textcolor{blue}{
Hence the synthesis times are set to the same number $N$ for each document cluster $\mathcal{C}_j$. 
}
This could be changed if we have more prior information regarding the distribution of different search intents.

When iterating on the $j^{th}$ cluster $\mathcal{C}_j$, the click-based relevance score for the $i^{th}$ document, denoted as $r_i^{c,Sc}$, is synthesized following the Bernoulli distribution with a constraint of their sum is $n_h$:

\begin{equation}
	  \forall  i,  r_ {i}^ {c}, Sc  \sim   \begin{cases}
	  Bernoulli(p_{c,rel}), \quad \text{if}\quad
  d_ {i}  \in  \mathcal{C}_ {j}
  \\Bernoulli(p_{c,irel}),\quad \text{if}\quad
  d_{i}  \notin  \mathcal{C}_{j}
  \end{cases}  
  \text{subject to}  \sum_{i}^{h}{r_i^{c,sc}}  =  Ci_{h}  
\end{equation}

where $p_{c,rel}$~($p_{c,irrel}$) is a parameter inferred from the distribution of the brain-based relevance scores in the user study's relevant~(irrelevant) documents with an interval estimation, i.e., $p_{c,rel}$ ($p_{c,irrel}$) indicates the possibility that a relevant~(irrelevant) document is clicked. 
The synthesis ensures that each $r_i^{c,Sc}$ follows the Bernoulli distribution, and the synthesized total number of clicks in $\mathcal{D}_h$ is $n_h$. 
This constraint is due to the number of total clicks being $n_h$ in a given search scenario $Sc$. To solve this constrained distribution in practice, we simply keep synthesizing a series of $r_i^{c,Sc}$~($i \in \{1, \ldots, h\}$) following the Bernoulli distributions until their sum is $n_h$.

On the other hand, the $i^{th}$ document $d_i$'s brain-based relevance score $r_i^{bs}$ is synthesized following a normal distribution:

\begin{equation}
	 r_ {i}^ {bs}  \sim  \begin{cases}Normal(\mu_{bs,rel},\sigma_{bs,rel}),\quad \text{if}\quad  d_ {i}  \in  \mathcal{C}_ {j} \\Normal(\mu_{bs,irel},\sigma_{bs,irel}),\quad \text{if} \quad d_ {i}  \notin  \mathcal{C}_ {j}\end{cases}   
\end{equation}
 
where $\mu_{bs,rel}$ and $\sigma_{bs,rel}$~($\mu_{bs,rel}$ and $\sigma_{bs,irel}$) are parameters inferred from the distribution of the brain-based relevance scores in the user study’s relevant~(irrelevant) documents with an interval estimation. 

\subsubsection{Optimal combination parameters searching.}  
After the synthesis process, for a search scenario $S_c = \{ q, \mathcal{D}_h, \mathcal{D}_u, n_h \}$, we synthesize $N$ corresponding $R_c$ and $R^{bs}$ for every document cluster $\mathcal{C}_j$~($j \in \{1, \ldots, q_m\}$). 
Then we search for the optimal combination parameters $\Theta^{it,Sc}$ and $\Theta^{re,Sc}$, respectively. 
For all selections of parameters, i.e., $\theta^{it,bs,Sc}$, $\theta^{it,c,Sc}$, and $\theta^{it,p,Sc}$ selected from $\{0, 0.2, 0.4, 0.6, 0.8, 1.0\}$, we compute the ranking-based evaluation metrics $\Pi(R^{gu}, R^{it})$, where $R^{gu}$ is evaluated by assuming documents belong to $\mathcal{C}_j$~($j \in \{1, \ldots, q_m\}$) as relevant:

\begin{equation}
r_ {i}^ {gu}  =  \begin{cases}1, \quad \text{if} \quad d_ {i}  \in  \mathcal{C}_{j} \\0,\quad \text{if} \quad d_ {i}  \notin  \mathcal{C}_{j}\end{cases}   
\end{equation}

Then we select the optimal combination parameters which achieve the averagely best RF performance~($\Pi$ is set as $NDCG@10$ in our experiment) among all $N \cdot q_m$ synthesized user signals $R^c$ and $R^{bs}$ for all document clusters $\mathcal{C}_j$~($j \in \{1, \ldots, q_m\}$). 
Note that we could not acknowledge what subtopic a search engine user is interested in advance when they submit the query. Hence, the optimal combination parameters that have average best performances for all subtopics are selected by the proposed method.

\input{meta/Table_5}

\input{meta/Table_6}

\subsubsection{Experimental Results}
Table~\ref{tab:para_irf} presents the experimental results of IRF using adaptive RF signals combination methods and fixed combination methods. 
From Table~\ref{tab:para_irf}, we observe that using adaptive RF combination methods is beneficial to IRF performance. 
This finding verifies our analyses and findings in Section~\ref{5.3}. 
Furthermore, we observe that the performance improvement brought by brain signals differs depending on whether applying the adaptive RF signals combination method or not. 
Specifically, the additional improvement brought by brain signals is 1.5\%~($QE^{F_{\theta^{it}}(R^{bs},R^c,R^p)}$ in comparison with $QE^{F_{\theta^{it}}(R^c,R^p)}$) in terms of NDCG@10 when using fixed combination parameters.
On the other hand, the performance difference of $QE^{F_{\theta}^{it,Sc}(R^{bs},R^c,R^p)}$ and $QE^{F_{\theta}^{it,Sc}(R^c,R^p)}$ is 8.8\% in terms of NDCG@10. This indicates that the adaptive RF signals combination method can further exploit the benefits brought by brain signals in the context of RF.

However, we observe that adopting the same adaptive RF signals combination method in RRF leads to no significant performance difference, as shown in Table~\ref{tab:para_rrf}. For example, the performance difference between $F_\Theta^{it,Sc}(R^{bs}, R^{c}, R^{p})$ and $F_\Theta^{it}(R^{bs}, R^{c}, R^{p})$ is not significant in terms of NDCG@10, where $\Theta^{it,Sc}$ is the adaptive combination parameter for RRF. 
This is due to the potential of the adaptive RF signals combination method is limited in RRF, as detailed in Section~\ref{5.3.1}. 
The exploration on how to better utilize multiple RF signals in RRF is left as future work.

Moreover, the experimental results show the possibility of improving RF performance by adaptively adjusting the combination weight of different RF signals. 
However, the proposed algorithm requires document clustering for each query. In our experiment, the selected queries are broad and sometimes ambiguous, and the document clusters are usually related to the user’s different search intents. 
Therefore, whether this algorithm is suitable for other types of real-world queries remains unknown. The exploration of designing more general algorithms to improve RF performance is left as future work.

\paragraph{Answer to \textbf{RQ4}.} 
We devise an adaptive RF signals combination method to re-weight the importance of various RF signals~(brain signals, click signals, pseudo-relevance signals) according to the search scenarios. 
We observe significant performance improvement with the proposed method. 
This verifies the analyses in Section~\ref{5.3} and illustrates that brain signals can boost IRF performance with an improvement of 8.8\% in terms of NDCG@10.

%% file: meta/algorithm_1.tex
\begin{algorithm}[t]
\caption{Adaptive RF Signals Combination}\label{algorithm}
\SetKwInput{KwInput}{Input} 
\SetKwInput{KwOutput}{Output} 
\KwInput{A search scenario $Sc = \{q,\mathcal{D}_h, \mathcal{D}_u, n_h\}$, where $n_h$ is selected from $\{1, 2, \ldots, h\}$; Documents clustering $\mathcal{D} = \{\mathcal{C}_1, \mathcal{C}_2, \ldots, \mathcal{C}_{qm}\}$. Synthesis times $T$.} 

\KwData{All candidate combination parameters $\Theta^{it} = \{\theta^{it,bs}, \theta^{it,c}, \theta^{it,p}\}$ where $\theta^{it,bs}$, $\theta^{it,c}$, $\theta^{it,p}$ are selected from $\{0, 0.2, 0.4, 0.6, 0.8, 1.0\}$; Best combination parameter $\Theta^{it,Sc}$ for search scenario $Sc$, initialized as $\{0, 0, 0\}$; Best synthesized performance $\hat{\Pi} = 0$.}

\KwOutput{The adaptive RF combination parameters $\Theta^{it,Sc}$.}

\For{each $\Theta^{it}$}{
    Sum of performance $\Pi = 0$;
    
    \For{each $\mathcal{C}_j \in \{\mathcal{C}_1, \mathcal{C}_2, \ldots, \mathcal{C}_{qm}\}$}{
        Synthesize $T$ possible click-based and brain-based relevance scores $\{R^{c,Sc}, R^{bs,Sc}\}$ in $Sc$ when assuming documents in $\mathcal{C}_j$ as relevant.
        
        Calculate the averaged RF performance $\Pi_{\mathcal{C}_j}$ with the synthesized relevance scores $\{R^{c,Sc}, R^{bs,Sc}\}$ and candidate combination parameter $\Theta^{it}$.
        
        $\Pi = \Pi + \Pi_{C_j}$;
    }
    \If{$\Pi > \hat{\Pi}$}{
        $\Theta^{it,Sc} = \Theta^{it}$;
        $\hat{\Pi} = \Pi$;
    }
}
\textbf{Return} $\Theta^{it,Sc}$;
\end{algorithm}

%% file: meta/Table_5.tex
\begin{table}
\caption{The document re-ranking performance in IRF, where $\Theta^{it}$ and $\Theta^{it,Sc}$ indicate fixed and adaptable combination parameters, respectively. $*$ indicates a significant performance difference when comparing $QE^{F_{\Theta^{it}}(R^{bs}, R^c, R^p)}$ to $QE^{F_{\Theta^{it,Sc}}(R^{bs}, R^c, R^p)}$, with a significance level of $p < 1 \times 10^{-3}$.\label{tab:para_irf}}
\setlength{\tabcolsep}{3mm}{
\begin{tabular}{@{}l!{\color{lightgray}\vrule}ccccc}
\specialrule{0em}{1pt}{1pt}
\toprule
\textbf{Method~\footnotemark[1]} & \textbf{NDCG@1} & \textbf{NDCG@3} & \textbf{NDCG@5} & \textbf{NDCG@10} & \textbf{MAP} \\ \midrule
$QE^{F_{\Theta^{it}}(R^c, R^p)}$ & 0.2842$^*$ & 0.2952$^*$ & 0.3124$^*$ & 0.3690$^*$ & 0.3708$^*$  \\
$QE^{F_{\Theta^{it}}(R^{bs}, R^c, R^p)}$ & 0.3056$^*$ & 0.3124$^*$ & 0.3285$^*$ & 0.3845$^*$ & 0.3826$^*$ \\ 
$QE^{F_{\Theta^{it,Sc}}(R^c, R^p)}$ & 0.2948$^*$ & 0.3024$^*$ & 0.3191$^*$ & 0.3747$^*$ & 0.3744$^*$ \\
$QE^{F_{\Theta^{it,Sc}}(R^{bs}, R^c, R^p)}$ & \textbf{0.3126} & \textbf{0.3258} & \textbf{0.3505} & \textbf{0.4183} & \textbf{0.4061}  \\ \bottomrule
\end{tabular}
}
\footnotetext[1]{$R^*$ indicates relevance score based on signals $^*$. $bs$, $c$, and $p$ indicate brain signals, click signals, and pseudo-relevance signals, respectively.}
\end{table}

%% file: meta/Table_6.tex
\begin{table}
\caption{The document re-ranking performance in RRF, where $\Theta^{it}$ and $\Theta^{it,Sc}$ indicate fixed and adaptable combination parameters, respectively. \textcolor{blue}{The utilization of adaptable combination parameters $\Theta^{re,Sc}$ instead of fixed combination parameters $\Theta^{re}$ does not yield any significant differences.}\label{tab:para_rrf}}
\setlength{\tabcolsep}{3mm}{
\begin{tabular}{@{}l!{\color{lightgray}\vrule}ccccc}
\specialrule{0em}{1pt}{1pt}
\toprule
\textbf{Method~\footnotemark[1]} & \textbf{NDCG@1} & \textbf{NDCG@3} & \textbf{NDCG@5} & \textbf{NDCG@10} & \textbf{MAP} \\ \midrule
$QE^{F_{\Theta^{it}}(R^{bs}, R^c, R^p)}$ &\textbf{ 0.6350} & 0.6617 & \textbf{0.7171} & 0.7693 & \textbf{0.8009}\\
$QE^{F_{\Theta^{it,Sc}}(R^{bs}, R^c, R^p)}$ & \textbf{0.6350} & \textbf{0.6622} & 0.7170 & \textbf{0.7694} & 0.8004  \\ \bottomrule
\end{tabular}
}
\footnotetext[1]{$R^*$ indicates relevance score based on signals $^*$. $bs$, $c$, and $p$ indicate brain signals, click signals, and pseudo-relevance signals, respectively.}
\end{table}

%% file: 6_conclusions.tex
\section{CONCLUSIONS AND DISCUSSIONS}
\label{6}
\subsection{Summary of Contributions}
In this paper, we propose a novel RF framework that combines pseudo-relevance signals, click signals, and brain signals for document re-ranking. 
Based on the proposed framework, we explore and verify the effectiveness of brain signals in the context of RF with different settings, i.e., IRF and RRF. 
In addition to analyzing the overall performance improvement brought by brain signals, we also dive into several search scenarios where click signals are missing or biased. 
We observe that brain signals are more helpful to IRF at the beginning of the search process before any clicks are received. 
We also demonstrate that brain signals can be applied to identify ``bad click'' and improve the performance of RRF. 
Besides, we further analyze how the RF performance varies with the combination weights across different search scenarios. We demonstrate that the importance of brain signals and click signals could vary in different scenarios. 
Based on this observation, we then propose to adaptively combine different RF relevance signals based on search scenarios, which leads to additional performance improvement.

\subsection{Discussions and Applications}
\label{6.2}
With the development of neurological devices, several applications have emerged to improve the performance of interactive information systems with BCI, e.g., personalized image editing~\cite{16davis2022brain}, crowdsourcing~\cite{17davis2020brainsourcing,19dikker2021crowdsourcing}. 
In IR, researchers also explore the possibility of establishing a pure BCI-based search system~(which does not require mouse or keyboard). 
\textcolor{blue}{
This opens up new avenues for understanding user intentions and enhancing search quality, while also highlighting potential challenges and concerns associated with brain-based relevance feedback~(RF).
}

\subsubsection{\textcolor{blue}{Privacy considerations}}
\textcolor{blue}{
The potential of leveraging brain responses in information systems is accompanied by significant privacy concerns, particularly regarding the possible abuse of data gathered from BCIs.
While the capabilities of anonymous technology in the field of brain signal processing are a cause for concern~\cite{bidgoly2022towards}, recent advancements have tried to mitigate this issue by edge computing, where users have the right to decide whether to upload data to the cloud. 
While our research doesn't delve into the intricacies of these technologies, it provides key insights into the privacy landscape: 
On the one hand, our findings underscore that non-deep brain decoding models with rapid inferencing ability can yield substantial results, thus accentuating the viability of edge computing. 
Additionally, even without user-specific data for model training, cross-subject models can adequately support RF, particularly in cold-start situations.
On the other hand, we observe that not every search scenario requires brain-based relevance feedback, but the importance of brain signals becomes paramount in specific search contexts. 
As such, users can opt to employ it selectively for tasks of higher difficulty.
}

\subsubsection{\textcolor{blue}{EEG availability}}
\textcolor{blue}{
Currently, wearable EEG devices are applied in tandem with VR technologies~\cite{tauscher2019immersive} and in services catering to the differently-abled~\cite{11chen2022web}. 
In IR-related research, functional information systems utilizing BCI have been developed. 
A notable instance is the work by \citet{11chen2022web}, who crafted a fully operational BCI-integrated search system that eliminates the need for traditional input devices such as keyboard and mouse. 
The decrease in cost and increase in portability of Brain-Computer Interfaces (BCIs) have been recognized as significant factors when applying BCI to IR research and daily applications.
Recently, \citet{maiseli2023brain} addresses the affordability and portability of BCIs, specifically mentioning the emergence of low-cost BCIs for everyday use, highlighting the progress towards making these technologies more accessible to the general public.
This research elucidates the ongoing efforts to reduce the cost and enhance the portability of BCIs, which in turn, is expected to significantly broaden the spectrum of their application scenarios, including in the field of IR.
What we add on top of existing BCI research is that we reveal another benefit of BCI for information systems, i.e., RF, besides directly controlling information systems. 
The closed loop between the user and the system can be built since the system can better understand the user with RF model.
}

\subsubsection{\textcolor{blue}{Comparison with conventional user signals}}
\textcolor{blue}{
Since traditional interaction paradigms can only capture indirect user feedback, biases in these behaviors have been long-standing IR research problems.
To address these challenges, we propose utilizing brain signals as a unique form of ``explicit feedback''. 
Concurrently, we identify specific search scenarios in which biases in clicks may adversely affect RF performance, highlighting search scenarios where brain signals can offer significant advantages.
}

\subsubsection{\textcolor{blue}{Recommended practices for EEG-based human study}}
\textcolor{blue}{
In our study, we aimed to implement settings that closely resemble real-world scenarios. 
For instance, instead of using random data splitting as done in prior research, we adopted a split-by-timepoint splitting to simulate the cold start situation that new users experience when they begin using our system. 
Furthermore, we conducted sensitivity analyses to demonstrate the impact of EEG pre-processing methods on relevance prediction performance.
Our study provides reliable evidence for the practical use of EEG signal processing in IR scenarios.
}

\subsubsection{\textcolor{blue}{Application Scenarios}}
\textcolor{blue}{
\textbf{Interactive IR:} We reveal the possibility of establishing more interactive IR systems with brain signals by devising a novel RF framework. 
This gives insights into existing BCI-based search systems, in which users can not only search with their thoughts~(without hand-based interactions) but also improve their search quality with a novel human-machine loop based on RF.
}

\textcolor{blue}{
\textbf{Human research for IR:} It has been widely recognized that RF is not only a potentially useful technique for improving search quality but also an effective tool to investigate how people search~\cite{59ruthven2003survey}. 
The combined relevance score can not only be utilized for document re-ranking~(IRF and RRF), but also help us understand how people actually interact with systems. 
Especially in search scenarios in which conventional RF signals may be biased, where brain signals are verified as a helpful substitute to infer relevance.
}

\textcolor{blue}{
\textbf{Human-Computer Interaction~(HCI) applications:}
In addition to Web search scenarios, our relevance feedback framework can provide inspiration to several vertical applications related to information retrieval. 
For example, a conversational chatbot may benefit from RF, especially in cases where users have difficulty expressing their thoughts in plain text but have ideas about the returned content~\cite{34kiesel2021meant}. 
Note that our proposed method for combining various RF signals is not restricted to combining brain signals with click signals and pseudo-relevance signals in Web search scenarios. 
Hence this RF method can adapt to combination with other types of user signals and broader search applications and HCI products.
}

\subsection{Limitations \& Future Work} 
\label{6.3}
We evaluate the feasibility of combining brain signals and conventional feedback signals for two typical RF tasks, i.e., IRF and RRF, for the first time. 
The experiments and analyses not only demonstrate that brain signals are feasible sources for RF, but also illustrate several search scenarios in which brain signals are more helpful to existing RF. 
Nonetheless, our research is not without its limitations, which may guide future work as follows:

\subsubsection{\textcolor{blue}{Selection of search tasks}}
\textcolor{blue}{
We derived our search tasks from TREC and iMine, characterized by their typically short and broad-topic queries. 
Given that RF doesn't universally benefit all queries and search scenarios~\cite{4azad2019query}, we prioritized these queries due to their heightened potential for RF advancements compared to more specific ones. 
This makes them a reasonable starting point to investigate the impact of brain signals on such standard queries. 
Future work may delve into a wider variety of query types and their synergy with brain signals in RF. 
Additionally, our approach extends beyond conventional query-based searches. 
The potential of our proposed RF framework also lies in its applicability to emerging search systems, like conversational systems, where traditional documents transition into diverse system response formats.
}

\subsubsection{\textcolor{blue}{User study settings}}
\textcolor{blue}{
First, our user study adopted a relatively strict design to mitigate potential confounders affecting brain signal collection.
For example, complex search behaviors in real life such as revisit~\cite{67xu2012incorporating}, and comparison~\cite{77zhang2021constructing} are not considered in our study to avoid uncontrolled bias. 
This constraint is not unique to our approach but is inherent in any behavioral signals~\cite{5azzopardi2021cognitive} used for search performance evaluation. Nevertheless, our work pioneers the integration of brain signals in RF within web search contexts. Contrasted with existing NeuraSearch literature~\cite{49moshfeghi2013understanding,56pinkosova2020cortical}, our study employs more authentic queries and Web pages sourced from prevalent commercial search engines. The challenge of navigating intricate scenarios with unfettered user behaviors remains a prospect for subsequent studies.
}

\subsubsection{\textcolor{blue}{Comparation with more user signals}}
\textcolor{blue}{
In information retrieval research, apart from click signals and brain signals, there are also user signals such as eye-tracking and mouse movements.
However, existing research has shown that eye-tracking and mouse movements are not accurate enough since they are also indirect probes of user signals~\cite{mao2014estimating}. 
The main difference between brain signals and conventional signals is that brain signals provide direct feedback on a person’s consciousness and thoughts.
The only concern lies in the noise introduced during brain signal collection and decoding, which can be further mitigated through improved equipment and data quality. 
In our study, we only compare brain signals with pseudo-relevance signals and click signals since they are the most commonly used signals in RF. 
However, the idea of integrating eye movement data and more modalities in a future investigation is intriguing and holds merit.
}


%% file: 7_acknowledgement.tex
\section{ACKNOWLEDGEMENT}
This work is supported by Quan Cheng Laboratory (Grant No. QCLZD202301).